\documentclass[10pt,twocolumn,letterpaper]{article}
\usepackage[pagenumbers]{cvpr} 

\usepackage[ruled,vlined]{algorithm2e}
\usepackage{siunitx}
\usepackage{multirow}
\usepackage{float}
\usepackage{placeins}
\usepackage{makecell}
\usepackage{pifont}
\usepackage[accsupp]{axessibility} 

\usepackage{bibunits}
\defaultbibliographystyle{ieeenat_fullname}

\definecolor{cvprblue}{rgb}{0.21,0.49,0.74}
\usepackage[breaklinks,colorlinks,allcolors=cvprblue]{hyperref}
\def\confName{CVPR}
\def\confYear{2026}
\title{E2EGS: Event-to-Edge Gaussian Splatting for Pose-Free 3D Reconstruction}
\author{Yunsoo Kim, Changki Sung, Dasol Hong, Hyun Myung\thanks{Corresponding author}\\
Urban Robotics Lab, School of Electrical Engineering, KAIST\\
{\tt\small \{dbstn1121, cs1032, ds.hong, hmyung\}@kaist.ac.kr}
}
\begin{document}
\begin{bibunit}
\maketitle
\begin{abstract}
The emergence of neural radiance fields (NeRF) and 3D Gaussian splatting (3DGS) has advanced novel view synthesis (NVS). These methods, however, require high-quality RGB inputs and accurate corresponding poses, limiting robustness under real-world conditions such as fast camera motion or adverse lighting. Event cameras, which capture brightness changes at each pixel with high temporal resolution and wide dynamic range, enable precise sensing of dynamic scenes and offer a promising solution. However, existing event-based NVS methods either assume known poses or rely on depth estimation models that are bounded by their initial observations, failing to generalize as the camera traverses previously unseen regions. We present E2EGS, a pose-free framework operating solely on event streams. Our key insight is that edge information provides rich structural cues essential for accurate trajectory estimation and high-quality NVS. To extract edges from noisy event streams, we exploit the distinct spatio-temporal characteristics of edges and non-edge regions. The event camera's movement induces consistent events along edges, while non-edge regions produce sparse noise. We leverage this through a patch-based temporal coherence analysis that measures local variance to extract edges while robustly suppressing noise. The extracted edges guide structure-aware Gaussian initialization and enable edge-weighted losses throughout initialization, tracking, and bundle adjustment. Extensive experiments on both synthetic and real datasets demonstrate that E2EGS achieves superior reconstruction quality and trajectory accuracy, establishing a fully pose-free paradigm for event-based 3D reconstruction.
\end{abstract}    
\section{Introduction}
\label{sec:intro}

The advent of neural radiance fields (NeRF)~\cite{Mildenhall21} and 3D Gaussian splatting (3DGS)~\cite{Kerbl23} has revolutionized the field of novel view synthesis (NVS), enabling photorealistic rendering of complex 3D scenes from sparse input views. These volumetric representation methods typically take camera poses and 2D views as input, leveraging multi-view images to learn implicit or explicit 3D scene representations. Despite their remarkable success, these approaches fundamentally assume high-quality input images. This assumption makes them vulnerable to common real-world challenges, such as motion blur and adverse lighting conditions that frequently occur during rapid camera movements or in low-light environments.

Event cameras present a paradigmatic shift in visual sensing by capturing pixel-level brightness changes asynchronously rather than traditional frame-based acquisition~\cite{Gallego20}. They offer exceptional temporal resolution and high dynamic range, making them particularly effective for handling motion blur and adverse lighting conditions. The high temporal resolution of event cameras enables precise capture of rapid scene dynamics, motivating numerous studies in trajectory estimation~\cite{Hu22, Klenk24, Niu25} and event-based deblurring~\cite{Klenk23, Qi23, Yu24, Deguchi24}. Furthermore, due to their principle of detecting brightness changes, event cameras naturally generate dense responses at edges and texture boundaries, providing rich structural information about the scene geometry~\cite{Chamorro22, Rebecq18} for robust visual perception.

Recently, the integration of event streams with NeRF and 3DGS frameworks has been explored. However, most existing event-based 3D reconstruction methods~\cite{Rudnev23, Klenk23, Low23, Yura25, Xiong24, Zahid25, Wu24} still require camera poses as input, limiting their applicability in scenarios where accurate pose estimation is challenging or unavailable. To address this limitation, IncEventGS~\cite{Huang25} was introduced as a pose-free approach that follows simultaneous localization and mapping (SLAM) principles. This method iteratively optimizes both camera poses and 3D Gaussians by comparing event streams with rendered images. While IncEventGS represents a significant step toward pose-free event-based 3D reconstruction, it relies heavily on depth estimation models~\cite{Ke2024} for 3D Gaussian initialization.

High-quality 3D reconstruction requires both accurate placement of 3D Gaussians and precise pose estimation. The importance of proper 3D Gaussian initialization has been established in previous studies~\cite{Huang25, Zahid25, Yura25}. Simultaneously, event cameras naturally encode rich structural information~\cite{Rebecq18, Chamorro22} that provides geometric constraints crucial for accurate pose estimation. However, existing methods fail to effectively leverage these two complementary aspects together. IncEventGS~\cite{Huang25} uses depth models for initialization but does not explicitly utilize the inherent characteristics of event streams to jointly optimize both Gaussian parameters and camera poses.

This fundamental gap introduces several limitations on the robustness and accuracy of current pose-free approaches. First, the system experiences substantial performance variation depending on depth availability, with significantly degraded results when depth models fail. Second, although depth information provides strong geometric constraints that assist pose estimation, depth models estimated from initial frames fail to generalize when cameras explore broader spatial regions beyond the initial coverage. This leads to accumulated trajectory errors that drastically degrade reconstruction quality as the sequence length increases. Lastly, uniform pixel-wise losses fail to distinguish between geometrically informative edges and noisy regions, causing sparse event noise to equally influence optimization and degrade both pose accuracy and reconstruction quality.

To overcome these limitations, we propose event-to-edge Gaussian splatting (E2EGS), a pose-free framework that leverages edge information derived solely from event streams. Our key insight is that event cameras inherently encode edge information. We explicitly extract and leverage this geometric structural information for both Gaussian optimization and pose estimation. We develop a noise-resistant edge detection method by analyzing temporal coherence patterns in event streams. By initializing Gaussians along detected edges and applying edge-weighted losses throughout optimization, our framework prioritizes geometric constraints over texture matching, enabling accurate pose estimation and high-quality reconstruction without depth supervision or pretrained models.

Our main contributions are threefold:
(1) A patch-based temporal coherence analysis that extracts noise-resistant edge maps from event streams.
(2) A geometry-centric 3D Gaussian learning framework with edge-aware Gaussian initialization and edge-weighted objectives.
(3) Comprehensive evaluations on synthetic and real datasets, demonstrating superior trajectory estimation accuracy and competitive reconstruction quality solely based on event stream.
\section{Related Work}
\label{sec:relatedworks}

\subsection{Event-based 3DGS}
Integrating event cameras with 3DGS has recently become a promising approach for robust NVS in challenging conditions. We categorize existing methods based on whether they rely on pre-obtained camera poses or jointly optimize poses during reconstruction.

\noindent\textbf{Methods with pre-obtained poses.}
Several works have explored event-based 3DGS under the assumption of known camera poses, typically obtained either from ground truth or structure-from-motion (SfM) methods~\cite{Hartley03, Snavely06, Ke2024} such as COLMAP~\cite{Schonberger16}. EventSplat~\cite{Yura25} and Elite-EvGS~\cite{Zhang25} use event-to-video models~\cite{Rebecq19} to generate pseudo-frames from event streams, then apply standard SfM pipelines for pose estimation and initial point cloud generation. EvaGaussians~\cite{Yu24} and E2GS~\cite{Deguchi24} merge event streams and blurred images to recover sharp scenes. They both depend on COLMAP~\cite{Schonberger16} for initial pose estimation, with EvaGaussians performing additional pose refinement using bundle adjustment. Ev-GS~\cite{Wu24} introduces the first event-supervised framework for 3DGS that optimizes 3D Gaussian parameters directly from event streams without RGB supervision. Event3DGS~\cite{Xiong24} deals with motion blur from egomotion in fast-moving robots by taking advantage of event cameras' high temporal resolution. EvGGS~\cite{Wang24} proposes a generalizable framework that jointly trains depth estimation, intensity reconstruction, and Gaussian regression, achieving cross-scene generalization. While these methods produce high-quality reconstructions, they require accurate external pose estimation, which restricts their use when poses are unavailable or in dynamic environments.

\begin{figure*}[!t]
    \centering
    \includegraphics[width=0.90\linewidth]{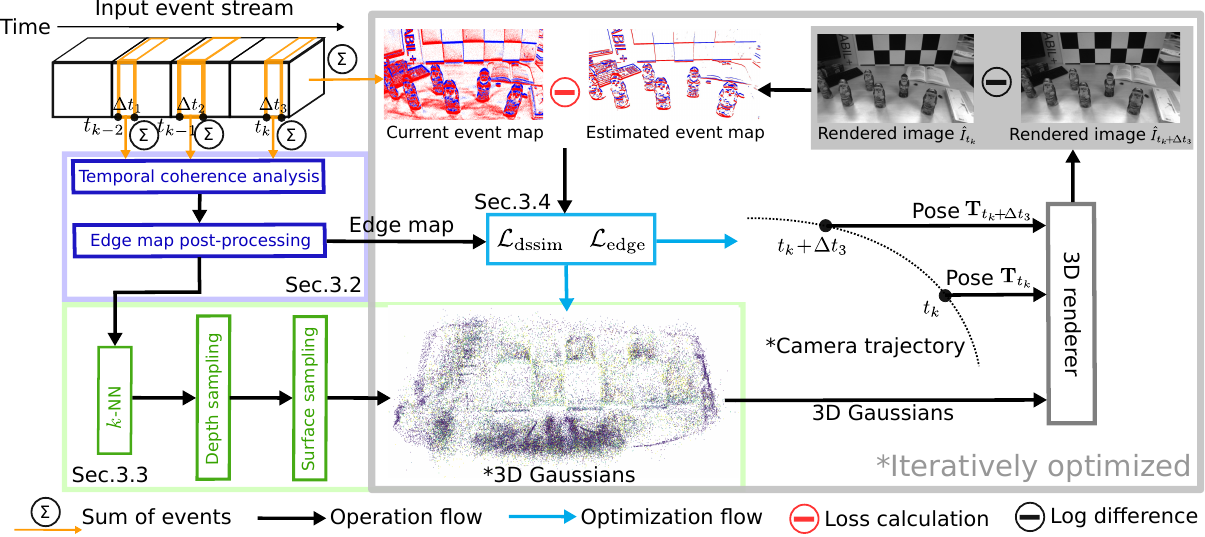}
    \vspace{-2mm}
    \caption{\textbf{Edge-guided reconstruction framework.} Our pipeline extracts robust edges from consecutive event maps (Sec.~\ref{sec:edge_detection}), initializes edge-aware Gaussians (Sec.~\ref{sec:edge_init}), and applies edge-guided losses during joint optimization of 3D scene representation and camera trajectory (Sec.~\ref{sec:edge_recon}). Depth sampling generates edge-guided Gaussians along viewing rays using inverse depth distribution, while surface sampling randomly initializes complementary Gaussians to cover texture-less regions. $k$-NN stands for $k$-nearest neighbors.}
    \label{fig:overview}
    \vspace{-4mm}
\end{figure*}

\noindent\textbf{Methods with pose optimization.}
Recent efforts aim to eliminate dependency on pre-computed poses, though most still rely on auxiliary modalities or pretrained models, limiting their autonomy. IncEventGS~\cite{Huang25} presents a SLAM-inspired framework that jointly optimizes camera poses and 3D Gaussians from event streams. However, it critically depends on pretrained depth estimation models~\cite{Ke2024} for initialization, raising questions about scalability and generalization to diverse scenes. GS-EVT~\cite{Liu25} formulates the task as cross-modal tracking, where event cameras localize against a pre-built 3D Gaussian map from frame-based cameras using differential image rendering. EGS-SLAM~\cite{Chen25} fuses RGB-D and event data for SLAM, explicitly modeling continuous camera trajectories to enable blur-aware tracking and high-fidelity reconstruction. Despite reducing direct pose dependency, these approaches still require external dependencies. Specifically, they rely on auxiliary modalities, such as RGB-D and blurred images, or on pretrained depth estimation models. However, such external dependencies limit their ability to operate independently across diverse scenarios. In contrast, our method achieves pose-free reconstruction from events alone, without relying on pretrained depth models or auxiliary sensors.

\subsection{Event-based visual odometry and SLAM}
Event cameras have strong potential for robust trajectory estimation due to their high temporal resolution and resilience to motion blur. Existing event-based odometry and SLAM methods can be categorized into feature-based, direct, and learning-based paradigms.

\noindent\textbf{Feature-based approaches.}
Line-based methods, such as event-based line SLAM~\cite{Chamorro22}, extract geometric primitives from event streams and establish correspondences for camera tracking using an error-state Kalman filter formulated on Lie manifolds. EVIT~\cite{Yuan24} presents event-based visual-inertial tracking in semi-dense maps using windowed nonlinear optimization. While these methods demonstrate the value of leveraging edge structures from event streams, they focus exclusively on pose estimation without reconstructing the underlying 3D scene for view synthesis. Our method extends this insight by embedding edge information directly into 3D Gaussian primitives, enabling unified optimization of trajectory and scene representations for both accurate pose estimation and high-quality NVS.

\noindent\textbf{Direct methods.}
Contrast maximization~\cite{Gallego18} has emerged as a powerful framework for event-based motion estimation, operating on the principle that correctly warped events produce sharp, high-contrast images. CMax-SLAM~\cite{Guo24} formulates camera pose estimation as a contrast maximization problem, optimizing motion parameters to maximize the sharpness of motion-compensated event images by aligning events along their actual trajectories. While they are effective for trajectory estimation, such methods focus on pose recovery without explicit 3D reconstruction for NVS.

\noindent\textbf{Learning-based approaches.}
Recent deep learning methods have achieved remarkable results. DEVO~\cite{Klenk24} introduces the first monocular event-only visual odometry system with strong real-world performance through deep patch selection and tracking. RAMP-VO~\cite{Pellerito24} presents an end-to-end learning framework that fuses asynchronous events and images via recurrent encoders. These methods demonstrate strong tracking performance but remain limited to odometry without scene reconstruction capabilities.
\section{Proposed Method}
\label{sec:methodology}

\subsection{Framework overview}
We adopt 3DGS~\cite{Kerbl23} as our scene representation, where each Gaussian primitive is parameterized by center position $\boldsymbol{\mu} \in \mathbb{R}^3$, covariance $\boldsymbol{\Sigma} \in \mathbb{R}^{3 \times 3}$, opacity $o$, and color $\boldsymbol{c}$. Given camera pose $\mathbf{T}$, images are rendered through differentiable $\alpha$-blending of projected 2D Gaussians.

Following IncEventGS~\cite{Huang25}, we process event streams in temporal chunks, each associated with a continuous trajectory parameterized by boundary poses $\mathbf{T}_{\text{start}}$ and $\mathbf{T}_{\text{end}} \in \text{SE}(3)$. The system alternates between tracking which estimates motion for new chunks, and mapping which performs joint bundle adjustment of trajectories and Gaussians within a sliding window. For event-based supervision, we accumulate event stream over the interval $\left[t,t+\Delta t \right]$ into event maps $E_t$ over the pixel domain $\Omega$, where $E_t(\mathbf{x})$ denotes the brightness change at pixel coordinate $\mathbf{x}\in\Omega$, and $E_t(P)$ denotes the event map restricted to region $P\subseteq\Omega$ at time $t$. $\Delta t$ is randomly sampled within $\left[\Delta t_\text{min}, \Delta t_\text{max}\right]$ to reflect the asynchronous nature of events. Event supervision minimizes the discrepancy between the measured event map $E_t(\mathbf{x})$ and synthesized event map $\hat{E_t}(\mathbf{x}) = \log \hat{I}_{t+\Delta t}(\mathbf{x}) - \log \hat{I}_t(\mathbf{x})$, where $\hat{I}_t$ denotes the brightness image rendered from 3DGS at time $t$.

\noindent\textbf{Edge-guided pipeline.}
Fig.~\ref{fig:overview} illustrates our edge-guided reconstruction framework. Given consecutive event maps from the input stream, we first extract robust edge maps through temporal coherence analysis (Sec.~\ref{sec:edge_detection}). These edge maps guide the initialization of 3D Gaussians, where edge-guided Gaussians are placed along detected edges with inverse log-depth sampling, complemented by random Gaussians for texture-less regions (Sec.~\ref{sec:edge_init}). During reconstruction, an edge-guided loss spatially weights the photometric error based on edge confidence, prioritizing optimization at geometrically salient boundaries (Sec.~\ref{sec:edge_recon}). This edge-aware initialization and optimization jointly refine the 3D Gaussian representation and camera trajectory, enabling robust pose estimation and high-quality reconstruction even in extended real-world sequences.

\subsection{Robust edge detection with patch-based temporal coherence analysis}
\label{sec:edge_detection}
We exploit the key insight that edges generate consistent temporal signatures across consecutive event maps. When the camera moves, edge produces correlated event responses detectable through temporal difference analysis. This approach is inspired by contrast maximization~\cite{Gallego18}, but rather than performing expensive trajectory estimation, we directly exploit temporal coherence to identify edge locations efficiently.

Given a sequence of $T$ consecutive event maps~$\{E_t\}_{t=1}^T$, we divide them into overlapping spatial patches of size ~\mbox{$p\times p$} with overlap ratio $\rho$. For each spatial patch location~$P_{x,y} = [x:x+p, y:y+p]$, we extract the temporal sequence of patches $\{E_t(P_{x,y})\}_{t=1}^T$ and compute pairwise temporal contrast between consecutive patches, measuring the consistency of edge responses.

For consecutive event maps, we compute the temporal difference signal with Gaussian windowing $G_\sigma$ to handle sharp temporal variations:
\begin{equation}
    D_t(P_{x,y}) = |G_\sigma \ast E_{t}(P_{x,y}) - G_\sigma \ast E_{t-1}(P_{x,y})|,
    \label{eq:temporal_diff}
\end{equation}
where $\ast$ denotes convolution.
As the camera moves, edges trigger events at spatially coherent locations across consecutive frames, producing structured spatial patterns in $D_t$ with high variance. In contrast, non-edge regions generate sparse, randomly distributed events with low variance. For each patch, we compute the contrast across all consecutive pairs and select the maximum:
\begin{equation}
    C(P_{x,y}) = \max_{t=2,...,T} \text{Var}(D_t(P_{x,y})),
\end{equation}
where patches with $C > \tau$ are classified as containing edges. We use an adaptive threshold $\tau$ to handle varying event statistics across different lighting conditions and camera motion speeds, which directly affect event frequency. For overlapping patches, we retain the maximum edge strength at each pixel location, forming a raw edge map $M_{\text{raw}}$. 

To obtain the final edge map $M$, we apply post-processing to $M_{\text{raw}}$ which includes Gaussian smoothing to reduce noise, adaptive thresholding to retain only strong edges, and morphological closing to connect nearby edge fragments. The resulting edge map $M \in [0,1]^{H \times W}$ encodes normalized edge confidence values at each pixel. Crucially, the temporal difference in Eq.~\ref{eq:temporal_diff} captures edges from both the previous frame $E_{t-1}$ and current frame $E_t$, enabling a self-correction mechanism. Edges with incorrectly estimated depth in previous frames can be identified and removed based on their inconsistency with current observations, ensuring only geometrically consistent edges guide subsequent Gaussian initialization and reconstruction.

\begin{table*}[t]
\caption{\textbf{NVS performance on Replica dataset.} Our method achieves superior reconstruction quality solely using event data. $^\dagger$\, denotes no depth supervision and $^*$ denotes that the method uses camera poses obtained through E2VID~\cite{Rebecq19} for frame reconstruction followed by COLMAP~\cite{Schonberger16} for pose estimation.}
\vspace{-2mm}
\centering
\resizebox{\textwidth}{!}{
\begin{tabular}{l|ccc|ccc|ccc|ccc|ccc}
\toprule
\multirow{2}{*}{Method} & \multicolumn{3}{c|}{room0} & \multicolumn{3}{c|}{room2} & \multicolumn{3}{c|}{office0} & \multicolumn{3}{c|}{office2} & \multicolumn{3}{c}{office3} \\
\cmidrule(lr){2-4} \cmidrule(lr){5-7} \cmidrule(lr){8-10} \cmidrule(lr){11-13} \cmidrule(lr){14-16}
 & PSNR↑ & SSIM↑ & LPIPS↓ & PSNR↑ & SSIM↑ & LPIPS↓ & PSNR↑ & SSIM↑ & LPIPS↓ & PSNR↑ & SSIM↑ & LPIPS↓ & PSNR↑ & SSIM↑ & LPIPS↓ \\
\midrule
EvGGS~\cite{Wang24} & {17.57} & {0.32} & {0.68} & {11.28} & {0.31} & {0.69} & {14.34} & {0.22} & {0.66} & {14.82} & {0.32} & {0.73} & {15.51} & {0.28} & {0.68} \\
Event-3DGS$^*$~\cite{Han24} & {22.27} & {0.84} & {0.30} & {16.39} & {0.68} & {0.36} & {15.97} & {0.64} & {0.51} & {18.19} & {0.60} & {0.48} & {17.82} & {0.84} & {0.31} \\
IncEventGS~\cite{Huang25} & \underline{23.54} & \underline{0.86} & \underline{0.20} & \textbf{23.25} & \underline{0.76} & \underline{0.27} & {26.53} & {0.49} & {0.44} & {21.27} & {0.76} & {0.29} & {19.21} & {0.82} & \underline{0.19} \\
IncEventGS$^\dagger$ & {19.81} & {0.83} & {0.24} & {22.58} & {0.73} & {0.29} & \underline{27.72} & \underline{0.51} & \textbf{0.40} & \underline{24.48} & \textbf{0.83} & \underline{0.26} & \underline{20.04} & \textbf{0.85} & \textbf{0.18} \\
\midrule
\textbf{Ours} & \textbf{23.86} & \textbf{0.87} & \textbf{0.19} & \underline{23.01} & \textbf{0.77} & \textbf{0.26} & \textbf{28.01} & \textbf{0.52} & \underline{0.41} & \textbf{24.86} & \textbf{0.83} & \textbf{0.25} & \textbf{20.75} & \textbf{0.85} & \underline{0.19} \\
\bottomrule
\end{tabular}
}
\label{tab:nvs_results}
\vspace{-4mm}
\end{table*}

\subsection{Edge-guided Gaussian initialization}
\label{sec:edge_init}

From the edge map $M$, we extract 2D edge points $\mathcal{P} =~\{p_i\}$. For each point, we identify its $k$-nearest edge neighbors and apply principal component analysis (PCA) to obtain the principal edge direction. The edge normal $\mathcal{N} = \{\textbf{n}_i\}$ is defined as the direction perpendicular to this tangent. We employ a recursive grid-based approach starting with tiles of size $s \times s$. For each tile, we compute the standard deviation of the edge normal orientations. If this angular deviation falls below threshold $\theta$, indicating consistent edge orientation, we create a single 2D Gaussian centered at the mean position of edge points within the tile, with orientation aligned to the average normal direction. Otherwise, we recursively subdivide the tile into four quadrants and repeat the process until convergence or the maximum depth $D$ is reached. This process yields a set of 2D edge Gaussians $\mathcal{G}_{\text{edge}} = \{g_i\}_{i=1}^{N_g}$, where $N_g$ denotes the total number of edge Gaussians.

Each 2D edge Gaussian in $\mathcal{G}_{\text{edge}}$ is extended to 3D through depth sampling along its viewing ray. We set the total number of 3D Gaussians to initialize as $N_{\text{total}}$, and introduce an edge ratio parameter $r_{\text{edge}}~\in~[0,1]$ that determines the proportion allocated to edge-guided 3D Gaussians: \mbox{$N_{\text{edge}}=\lfloor r_{\text{edge}}\cdot N_{\text{total}}\rfloor$}, with the remaining \mbox{$N_{\text{random}} = N_{\text{total}} - N_{\text{edge}}$} points randomly initialized for smooth surfaces.

For depth sampling along each edge ray, we employ inverse depth sampling to allocate more samples at farther distances:
\begin{equation}
d = \frac{1}{\frac{1}{d_{\max}} + u\left(\frac{1}{d_{\min}} - \frac{1}{d_{\max}}\right)}, \quad u \sim \mathcal{U}(0,1),
\end{equation}
where $d_{\min}$ and $d_{\max}$ denote the minimum and maximum depth bounds, respectively, and $\mathcal{U}$ is a uniform distribution. This distribution is geometrically motivated, as distant points exhibit larger pixel displacements under camera rotation, making them more informative for joint depth-pose optimization. By concentrating samples at far distances, we enhance the observability of rotational motion, thereby improving pose estimation accuracy.

To ensure balanced point distribution, we determine the number of depth samples $n_d = \lfloor N_{\text{edge}} / N_g \rfloor$. For each edge Gaussian $g_i$ located at pixel $\mathbf{x}_i$, we sample $n_d$ depths $\{d_{i,j}\}_{j=1}^{n_d}$ according to Eq.~(3) and generate 3D points along the viewing ray as:
\begin{equation}
\mathbf{X}_{i,j} = d_{i,j} \cdot \mathbf{K}^{-1}[\mathbf{x}_i, 1]^T,
\end{equation}
where $\mathbf{K}$ is the camera's intrinsic matrix and $\mathbf{K}^{-1}[\mathbf{x}_i, 1]^T$ represents the normalized ray direction corresponding to pixel $\mathbf{x}_i$.

\subsection{Edge-guided 3D reconstruction}
\label{sec:edge_recon}

To leverage detected edges during reconstruction, we introduce an edge-guided loss that spatially weights the reconstruction error based on edge confidence. Edge-weighted loss over pixel domain $\Omega$ can be calculated with given edge map $M$, synthesized event accumulation $\hat{E}$, and ground truth event accumulation $E$ as follows:
\begin{equation}
    \mathcal{L}_{\text{edge}} = \frac{1}{|\Omega|} \sum_{\mathbf{x} \in \Omega} w(\mathbf{x}) \cdot \|\hat{E}(\mathbf{x}) - E(\mathbf{x})\|^2,
    \label{eq:edge_loss}
\end{equation}
where the spatial weighting is $w(\mathbf{x}) = 1 + \beta \cdot M(\mathbf{x})$, with $\beta$ controlling edge emphasis. The total loss combines edge-guided and structural terms:
\begin{equation}
    \mathcal{L}_{\text{total}} = (1-\lambda)\mathcal{L}_{\text{edge}} + \lambda\mathcal{L}_{\text{dssim}},
\end{equation}
where $\lambda$ is a hyperparameter and $\mathcal{L}_{\text{dssim}} = 1-\text{SSIM}(E, \hat{E})$ is the structural dissimilarity loss~\cite{Wang04}. This prioritizes optimization at geometrically salient boundaries, enabling faster convergence and more accurate depth estimation at object edges.

\section{Experiments}
\label{sec:experiments}

\begin{table*}[t]
\caption{\textbf{Absolute trajectory error (ATE).} RMSE (cm) across synthetic Replica and real TUM-VIE sequences. Our method achieves competitive performance solely using event data.}
\vspace{-2mm}
\centering
\resizebox{0.8\textwidth}{!}{ 
\begin{tabular}{l|ccccc|ccccc}
\toprule
\multirow{2}{*}{Method} & \multicolumn{5}{c|}{Synthetic (Replica)} & \multicolumn{5}{c}{Real-world (TUM-VIE)} \\
\cmidrule(lr){2-6} \cmidrule(lr){7-11}
 & room0 & room2 & office0 & office2 & office3 & 1d & 3d & 6dof & desk & desk2 \\
\midrule
DEVO~\cite{Klenk24} & 0.271 & 0.381 & 0.287 & 0.134 & 0.356 & \textbf{0.23} & \underline{1.00} & 1.82 & 0.55 & 0.75 \\
ESVO2~\cite{Niu25} & - & - & - & - & - & \underline{1.08} & {1.51} & {1.49} & {0.89} & {2.65} \\
\midrule
IncEventGS~\cite{Huang25} & \underline{0.051} & \underline{0.071} & \underline{0.085} & \textbf{0.041} & \textbf{0.058} & 2.19 & 1.62 & \underline{0.70} & \underline{0.15} & \underline{0.41} \\
IncEventGS$^\dagger$ & 6.817 & 0.446 & 0.698 & 0.513 & 1.379 & 2.58 & 4.48 & 8.24 & 0.78 & 96.59 \\
\midrule
\textbf{Ours} & \textbf{0.049} & \textbf{0.065} & \textbf{0.078} & \underline{0.070} & \underline{0.077} & {1.12} & \textbf{0.65} & \textbf{0.58} & \textbf{0.12} & \textbf{0.40} \\
\bottomrule
\end{tabular}}
\label{tab:ate_results}
\vspace{-4mm}
\end{table*}

\subsection{Experiment settings}
\textbf{Implementation details.}
All experiments are conducted on a server with an AMD Ryzen Threadripper PRO 3955WX processor and NVIDIA RTX A5000 GPU. We adopt the 3DGS implementation~\cite{Kerbl23} and the tracking-mapping pipeline from IncEventGS~\cite{Huang25} with standard hyperparameter settings. For detailed parameter settings, please refer to the supplementary material.\\
\noindent\textbf{Baselines.}
We compare our E2EGS with baselines to validate the effectiveness of our approach for both 3D reconstruction and trajectory estimation. For 3D reconstruction evaluation, IncEventGS~\cite{Huang25}, the current state-of-the-art pose-free event-based 3DGS method, is compared using two configurations. The first uses Marigold~\cite{Ke2024} monocular depth estimation to initialize 3D Gaussian parameters, denoted as IncEventGS. The second uses random Gaussian initialization, denoted as IncEventGS$^\dagger$. We also evaluate Event-3DGS~\cite{Han24}, which requires camera poses as input, and EvGGS~\cite{Wang24}. For Event-3DGS, we obtain camera poses using E2VID~\cite{Rebecq19} for image reconstruction followed by COLMAP~\cite{Schonberger16}. For trajectory estimation, we compare the proposed method with IncEventGS~\cite{Huang25}, DEVO~\cite{Klenk24}, and ESVO2~\cite{Niu25}, which are publicly available state-of-the-art event visual odometry methods. ESVO2 is evaluated only on real datasets.\\
\noindent\textbf{Evaluation metrics.}
For NVS, we report standard metrics including peak signal-to-noise ratio (PSNR), structural similarity index (SSIM), and learned perceptual image patch similarity (LPIPS). Following EventNeRF~\cite{Rudnev23}, we apply linear color transformation between predictions and ground truth before metric computation, as event-based methods lack absolute brightness supervision. For camera pose estimation, we use absolute trajectory error (ATE), which measures the root-mean-square error (RMSE) of the aligned trajectory after SE(3) alignment using the EVO toolbox~\cite{Grupp2017}.\\
\noindent\textbf{Datasets.}
For synthetic evaluation, we use IncEventGS's event data from Replica~\cite{replica19} scenes (room0, room2, office0, office2, and office3). For real-world evaluation, we use TUM-VIE~\cite{Klenk21} sequences (1d, 3d, 6dof, desk, and desk2).

\subsection{Quantitative evaluations}
\noindent\textbf{Novel view synthesis.}
Tab.~\ref{tab:nvs_results} shows reconstruction quality on synthetic scenes. While baseline methods show significant performance variation across scenes, our method maintains more stable performance. This stability stems from our edge-guided optimization, which prioritizes geometric structures over photometric appearance. Without edge guidance, photometric error from event noise uniformly affects 3D reconstruction, causing optimization process to receive unreliable gradient signals. In contrast, our edge-centric approach focuses on structural boundaries where geometric constraints remain robust despite noise, achieving consistent reconstruction quality across diverse scenes.

\begin{figure}[t]
\centering
\includegraphics[width=0.9\linewidth]{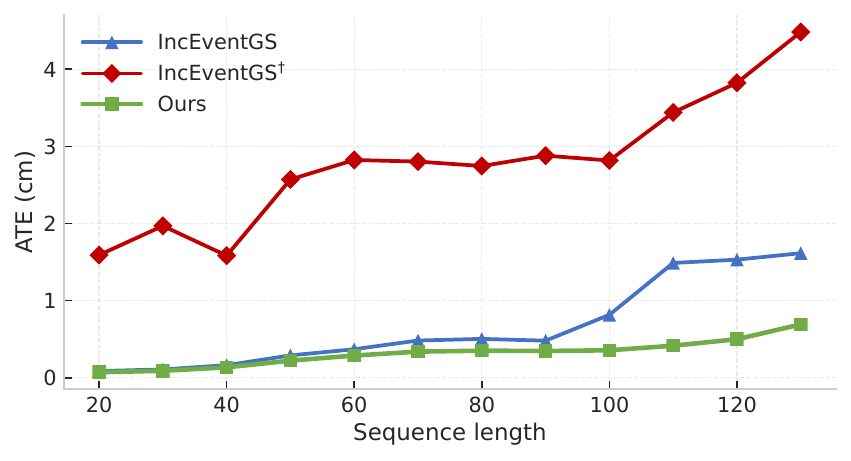}
\vspace{-4mm}
\caption{\textbf{ATE with respect to the length of the sequence.}}
\label{fig:ate_graph}
\vspace{-4mm}
\end{figure}

\noindent\textbf{Trajectory estimation.}
Tab.~\ref{tab:ate_results} shows pose estimation results. On synthetic Replica scenes, our edge-guided approach achieves sub-millimeter accuracy across all scenes. Without requiring any depth supervision, our edge-guided approach outperforms DEVO~\cite{Klenk24} and IncEventGS$^\dagger$ by substantial margins. Notably, even on synthetic datasets where event noise is minimal and depth estimation models achieve highly accurate results, our geometric edge constraints provide equally strong guidance for pose optimization.

On real-world TUM-VIE sequences, IncEventGS$^\dagger$ suffers from catastrophic failure due to the lack of geometric constraints in random initialization, causing pose optimization to converge to local minima. More interestingly, IncEventGS~\cite{Huang25} shows increased errors despite low errors on initial trajectory segments where camera motion remains within a small spatial region as shown in Fig.~\ref{fig:ate_graph}. This reveals that depth models, estimated from early frames, fail to generalize when cameras explore broader spatial regions beyond the initial coverage. In contrast, our edge-guided approach maintains consistently low errors across real-world sequences, demonstrating that edge-based priors provide more robust constraints than depth estimation for long-range trajectories.

\subsection{Qualitative evaluations}

\begin{figure*}[!ht]
\centering
\setlength{\tabcolsep}{1pt}
\footnotesize
\begin{tabular}{@{}lccc@{}}
& \textbf{room0} & \textbf{office0} & \textbf{office3}\\[0.5mm]
\rotatebox{90}{\hspace{3mm}Ground Truth} &
\includegraphics[width=0.39\linewidth]{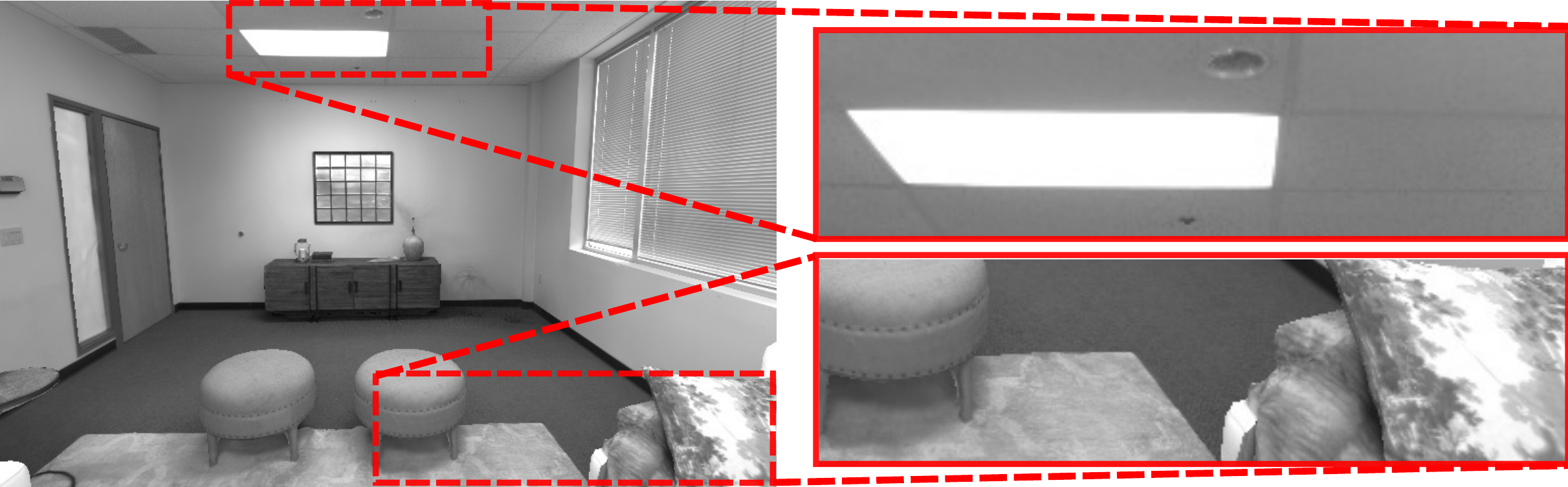} &
\includegraphics[width=0.33\linewidth]{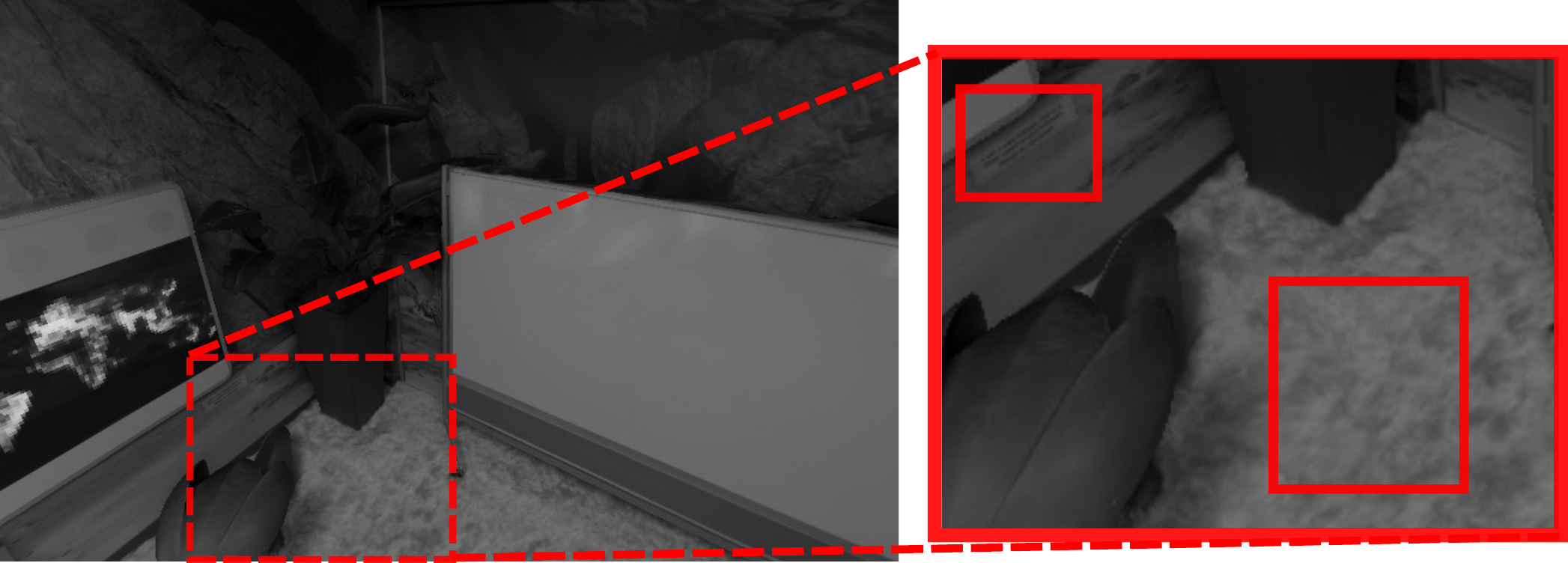} &
\includegraphics[width=0.19\linewidth]{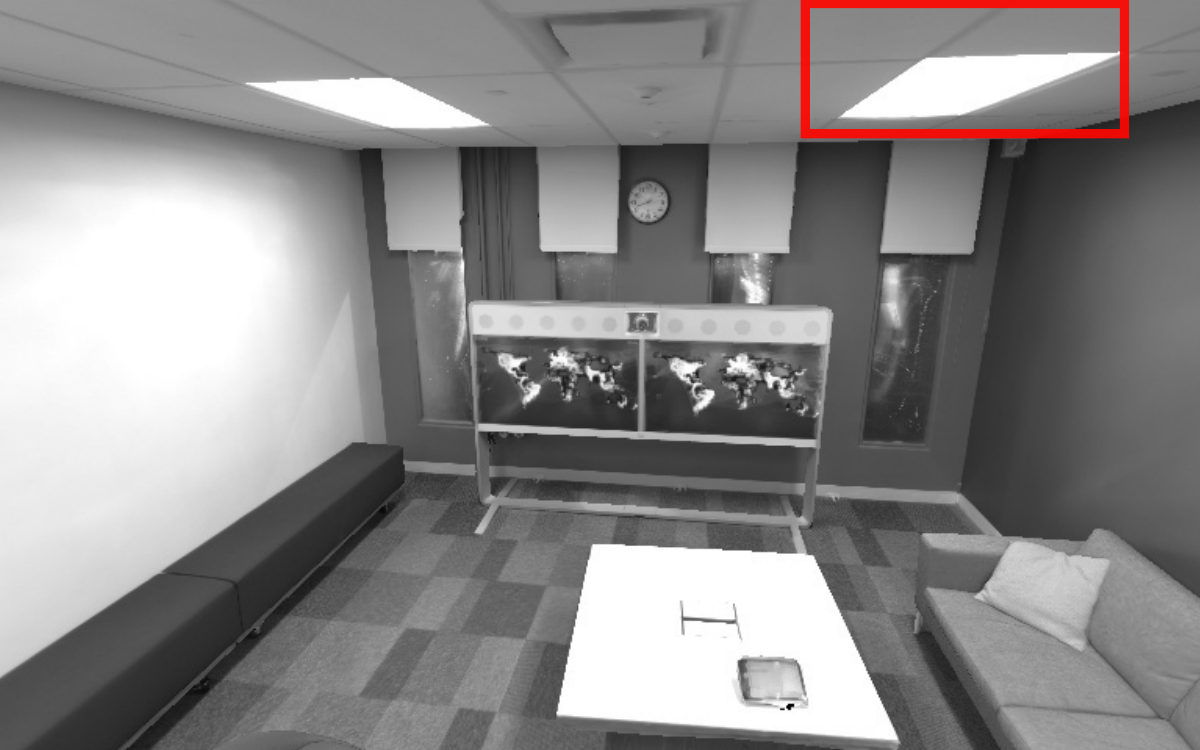} \\[0.1mm]
\rotatebox{90}{\hspace{1mm}IncEventGS~\cite{Huang25}} &
\includegraphics[width=0.39\linewidth]{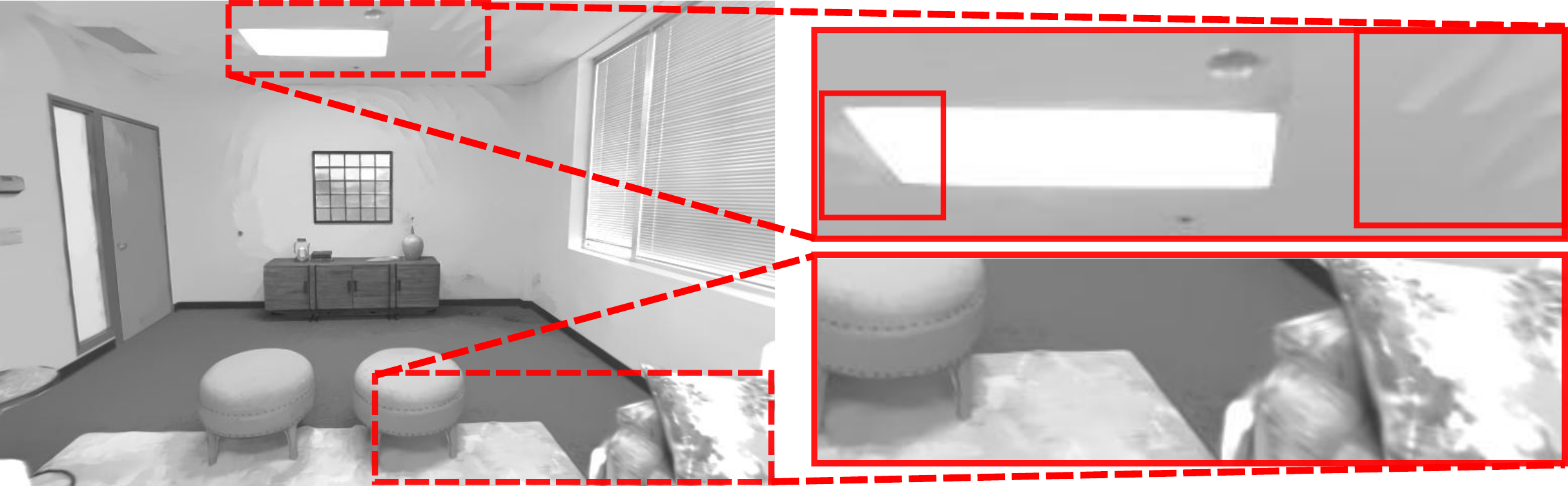} &
\includegraphics[width=0.33\linewidth]{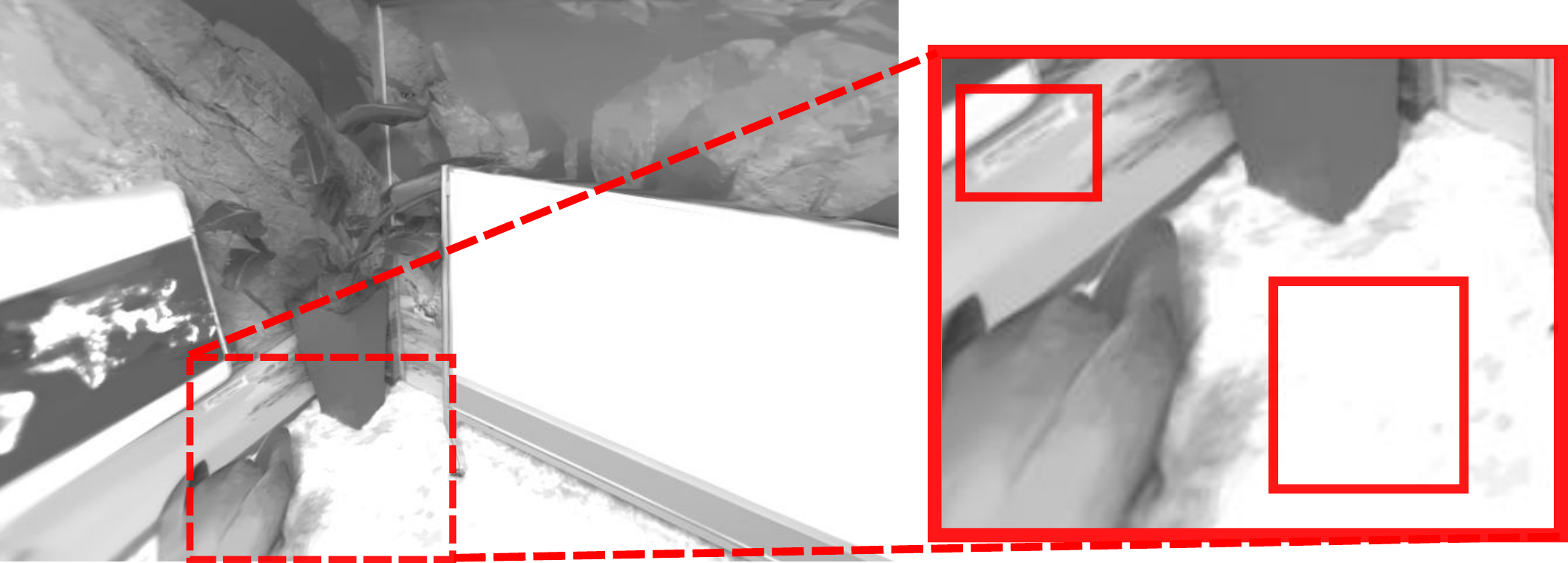} &
\includegraphics[width=0.19\linewidth]{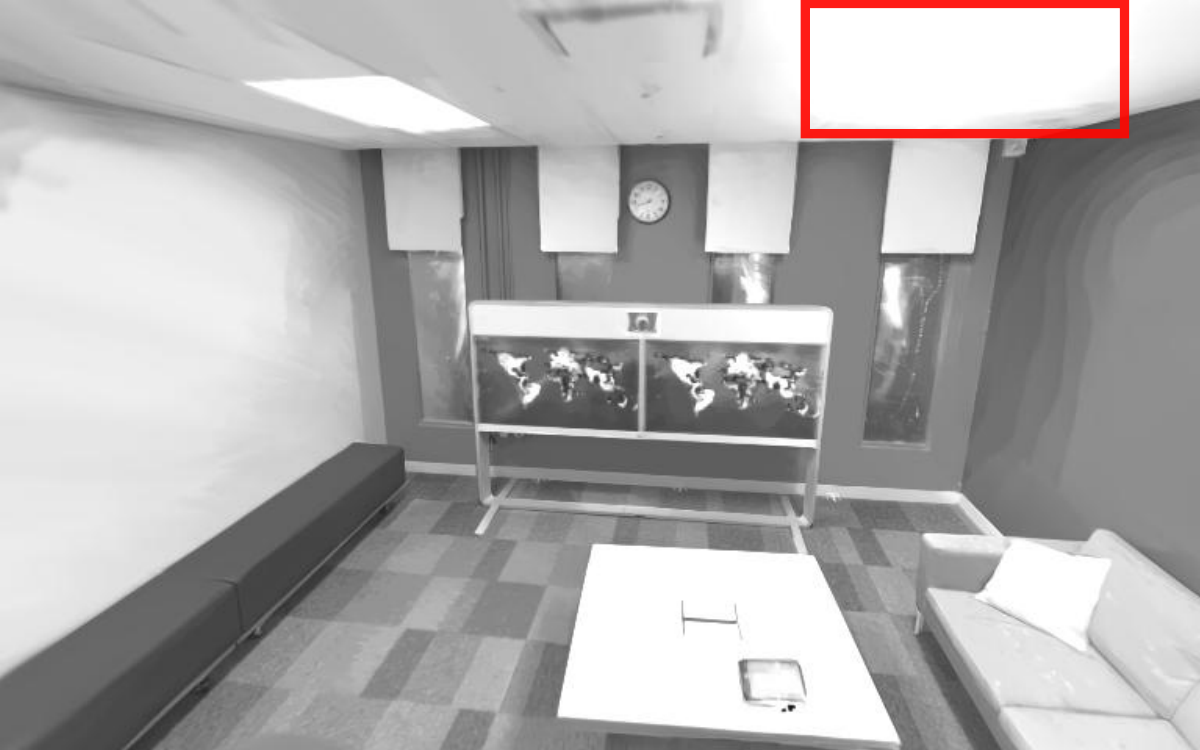} \\[0.1mm]
\rotatebox{90}{\hspace{3mm}IncEventGS$^\dagger$} &
\includegraphics[width=0.39\linewidth]{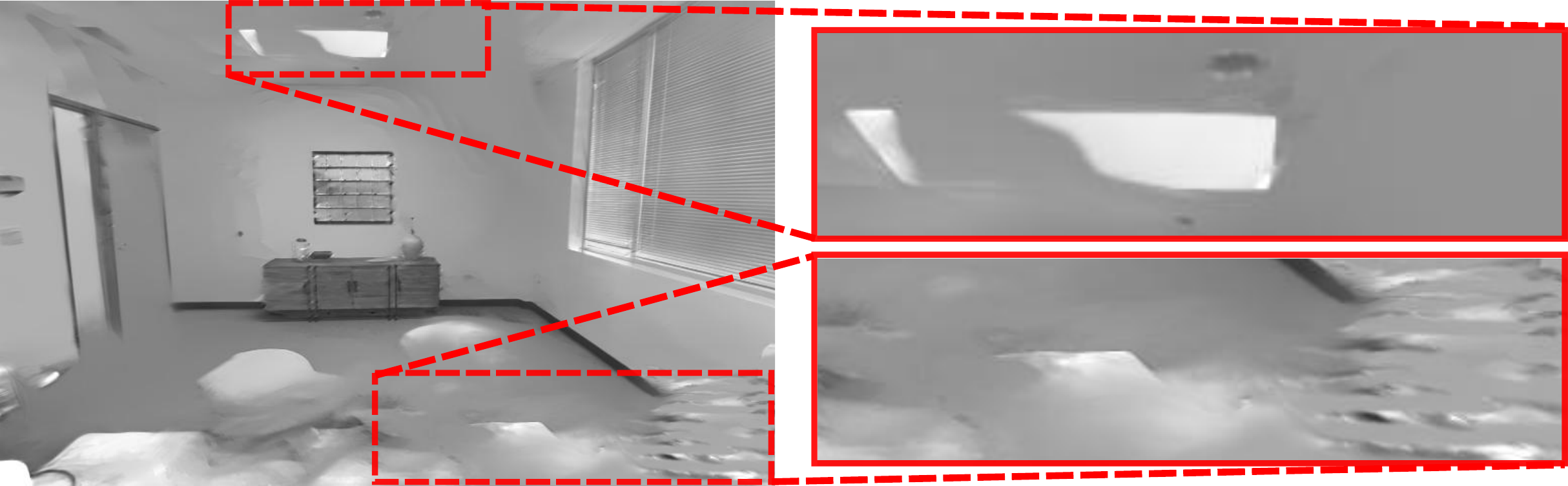} &
\includegraphics[width=0.33\linewidth]{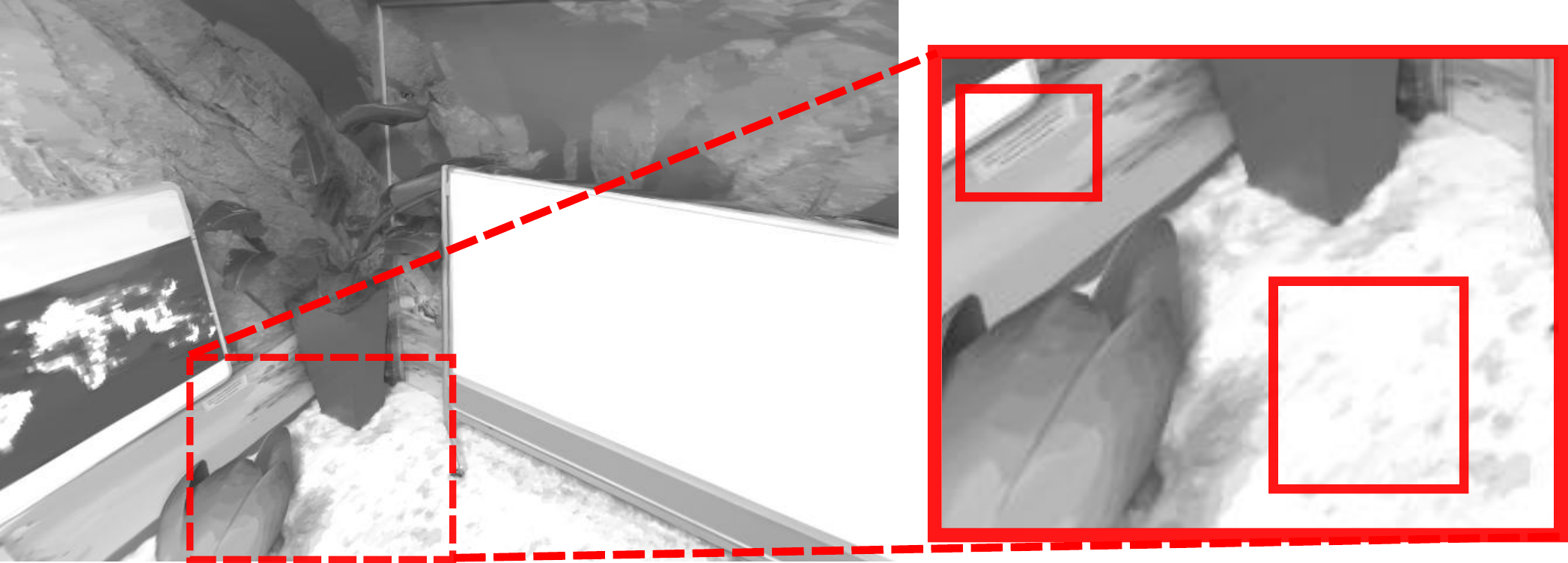} &
\includegraphics[width=0.19\linewidth]{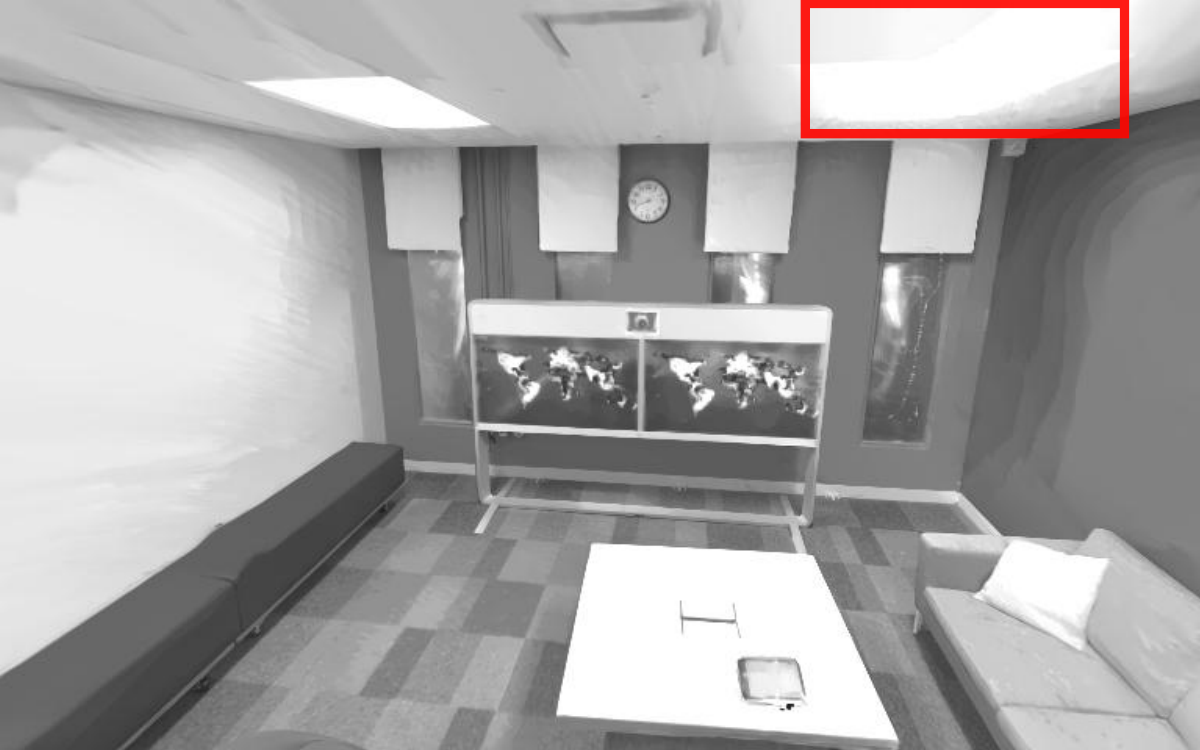} \\[0.1mm]
\rotatebox{90}{\hspace{8mm}Ours} &
\includegraphics[width=0.39\linewidth]{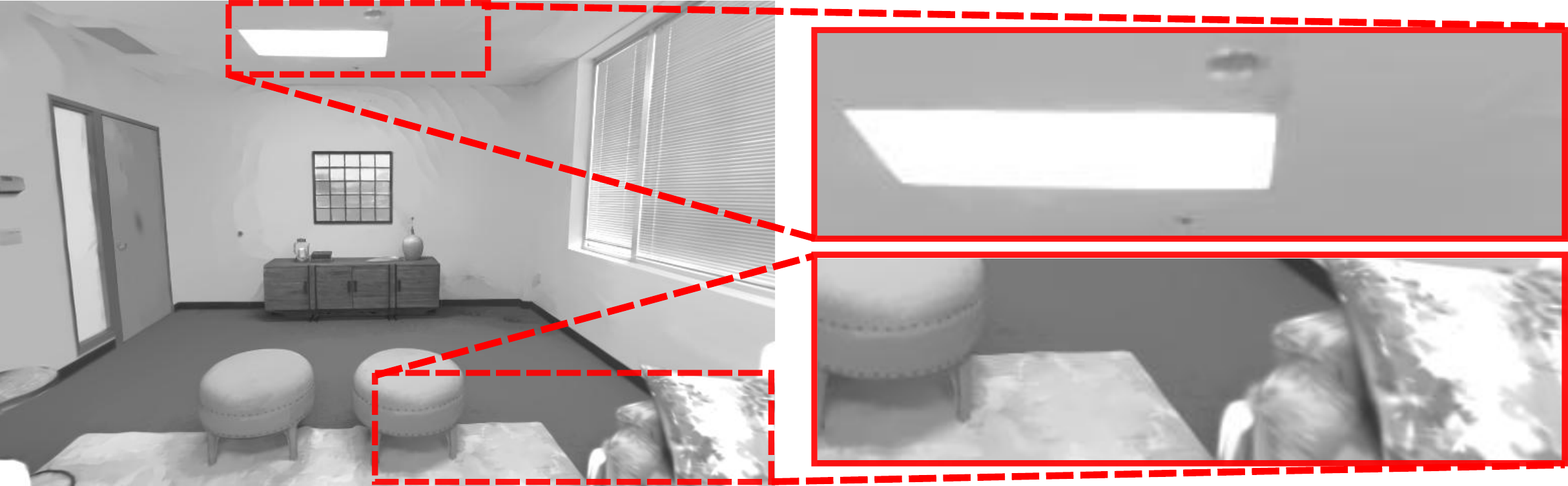} &
\includegraphics[width=0.33\linewidth]{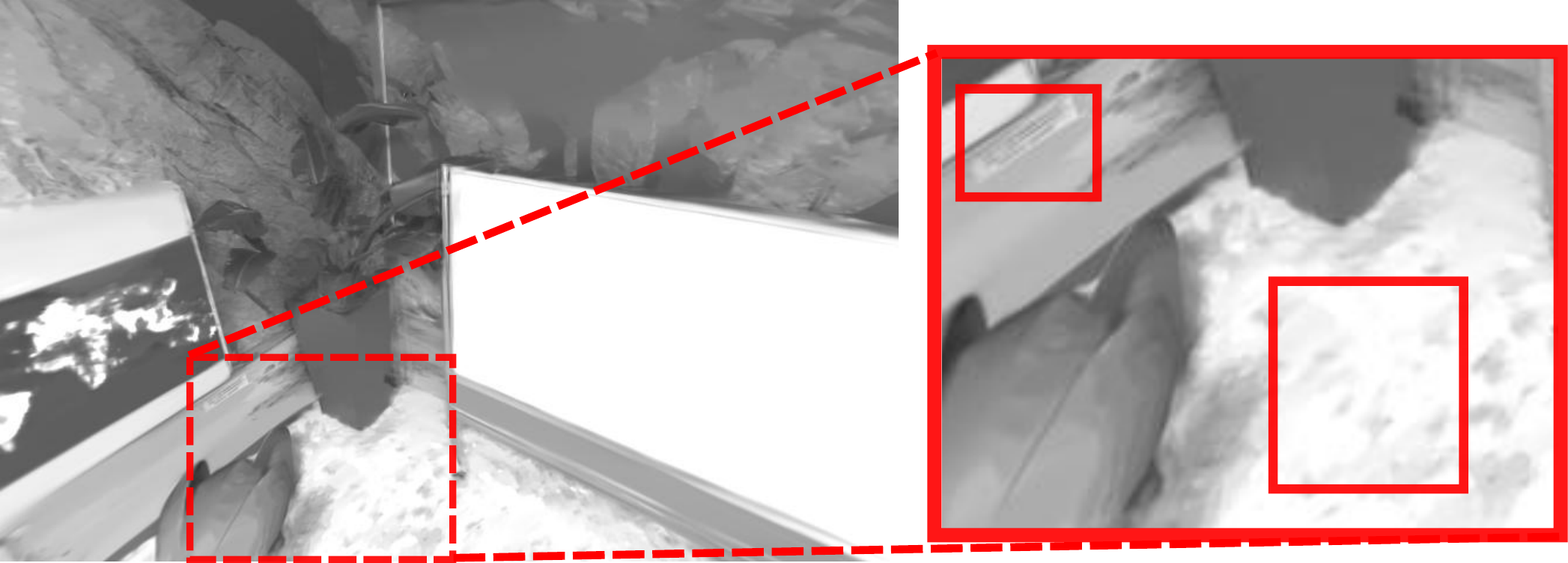} &
\includegraphics[width=0.19\linewidth]{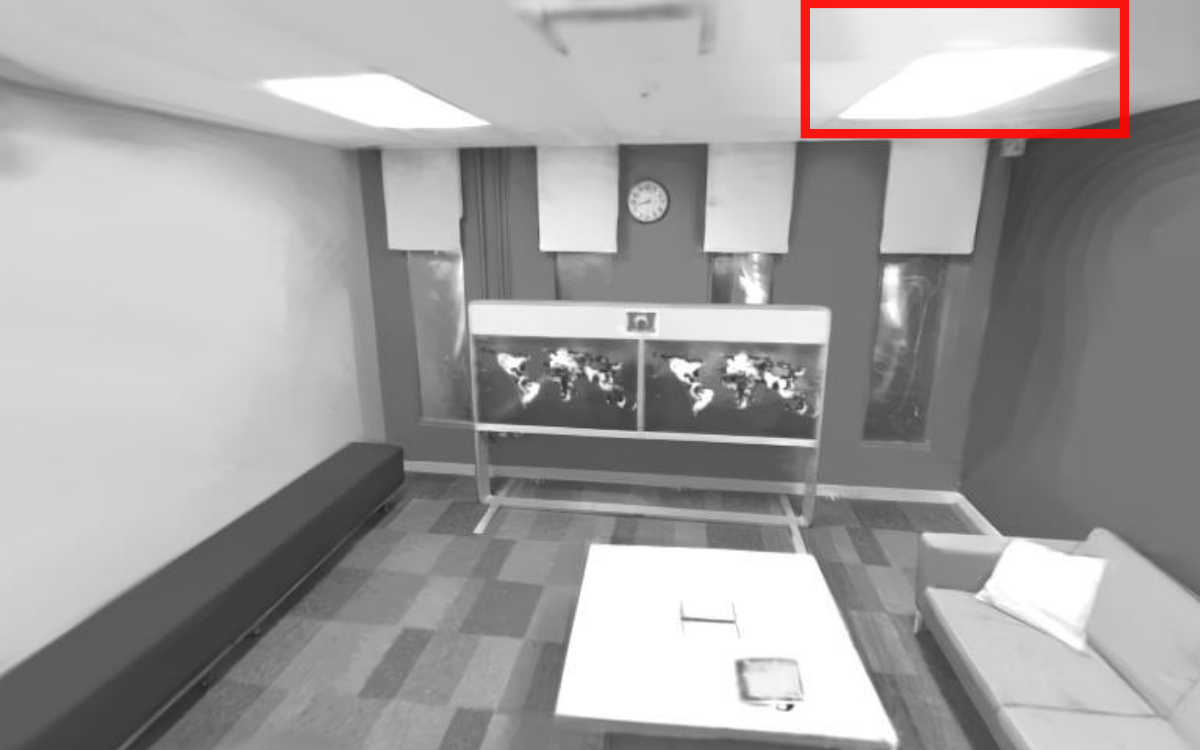}
\\
\end{tabular}
\vspace{-2mm}
\caption{\textbf{Qualitative results on Replica dataset.} 
Red boxes highlight regions of interest for comparison. Our method produces sharper boundaries and cleaner surfaces compared with baselines. IncEventGS shows failures including wave-like artifacts, missing details, and indistinct boundaries. IncEventGS$^\dagger$ exhibits severe reconstruction failures due to accumulated trajectory estimation errors.}
\label{fig:qualitative_replica}
\end{figure*}

\noindent\textbf{Novel view synthesis.}
Fig.~\ref{fig:qualitative_replica} presents reconstruction results across Replica scenes. IncEventGS exhibits various failure modes in regions highlighted by red boxes, including wave-like artifacts in texture-less regions, missing fine details such as textures and patterns, and indistinct object boundaries. These issues reveal the fundamental limitation of uniform photometric matching. In contrast, our edge-guided approach suppresses these artifacts by prioritizing geometric constraints over texture matching, achieving sharper boundaries and better detail preservation. IncEventGS$^\dagger$ produces severely degraded reconstruction due to accumulated trajectory errors from random initialization, demonstrating the importance of geometric priors for robust pose estimation.

\begin{figure*}[!t]
\centering
\setlength{\tabcolsep}{2pt}
\begin{tabular}{@{}cccc@{}}
\includegraphics[width=0.23\linewidth]{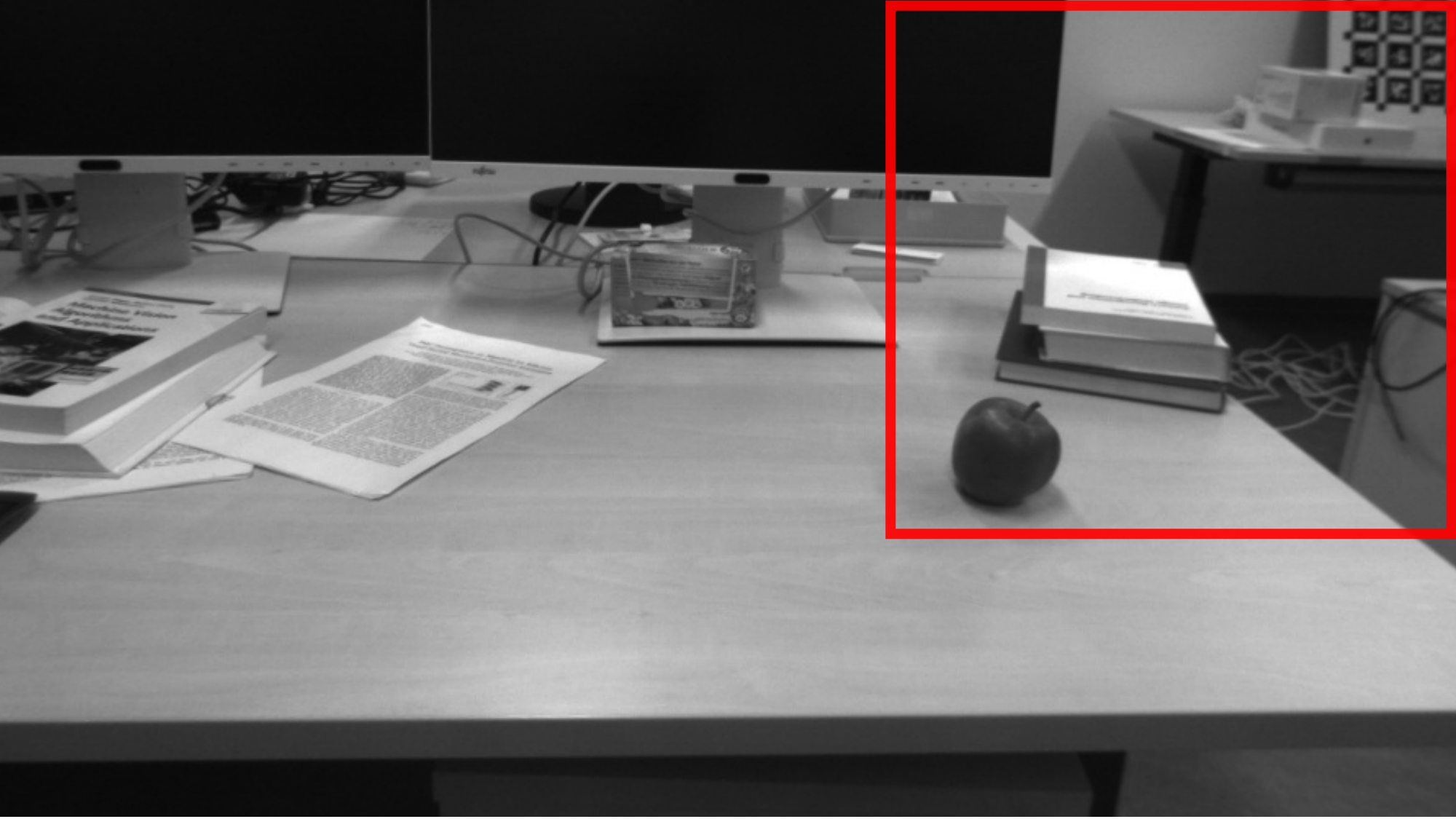} &
\includegraphics[width=0.23\linewidth]{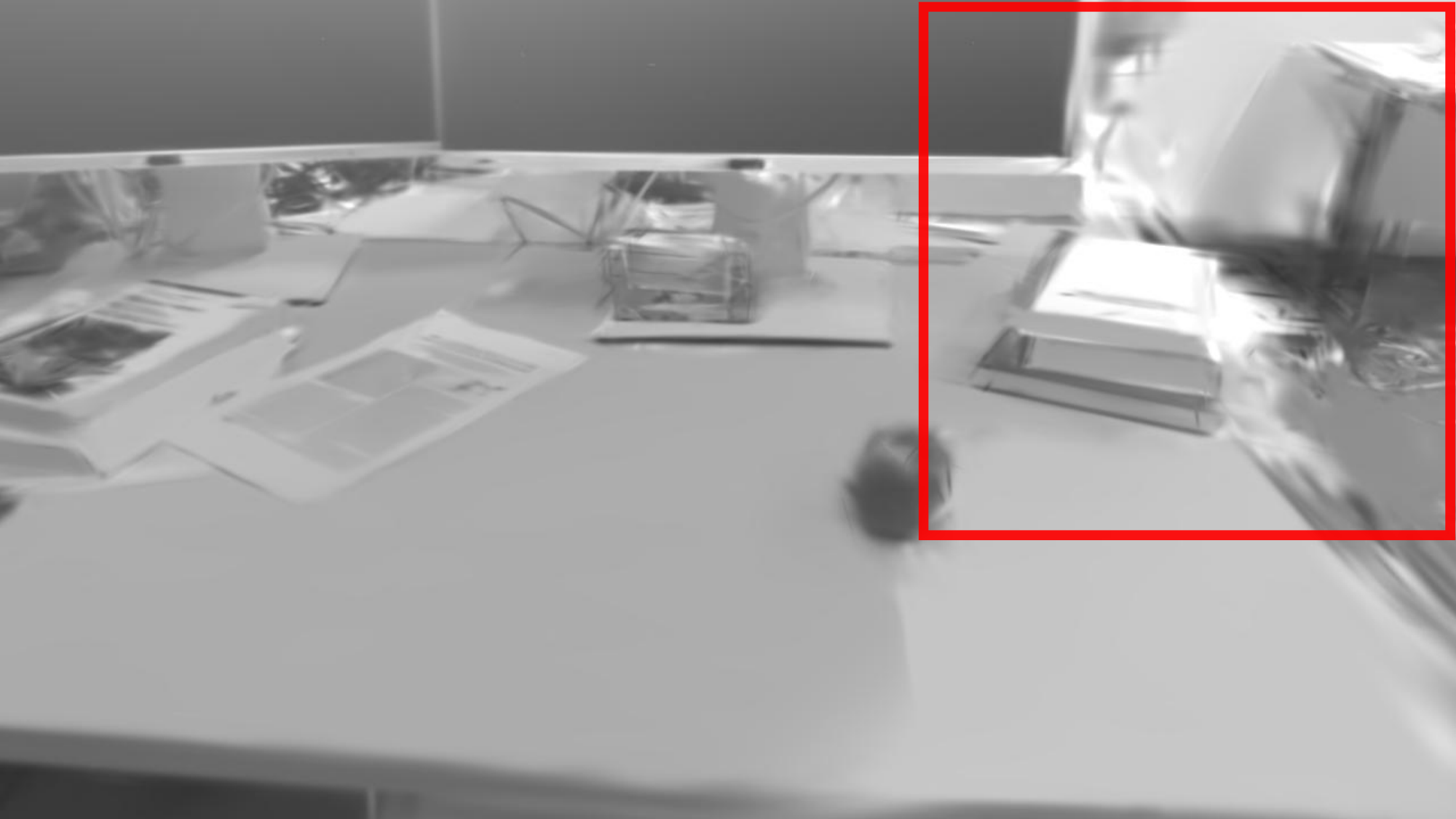} &
\includegraphics[width=0.23\linewidth]{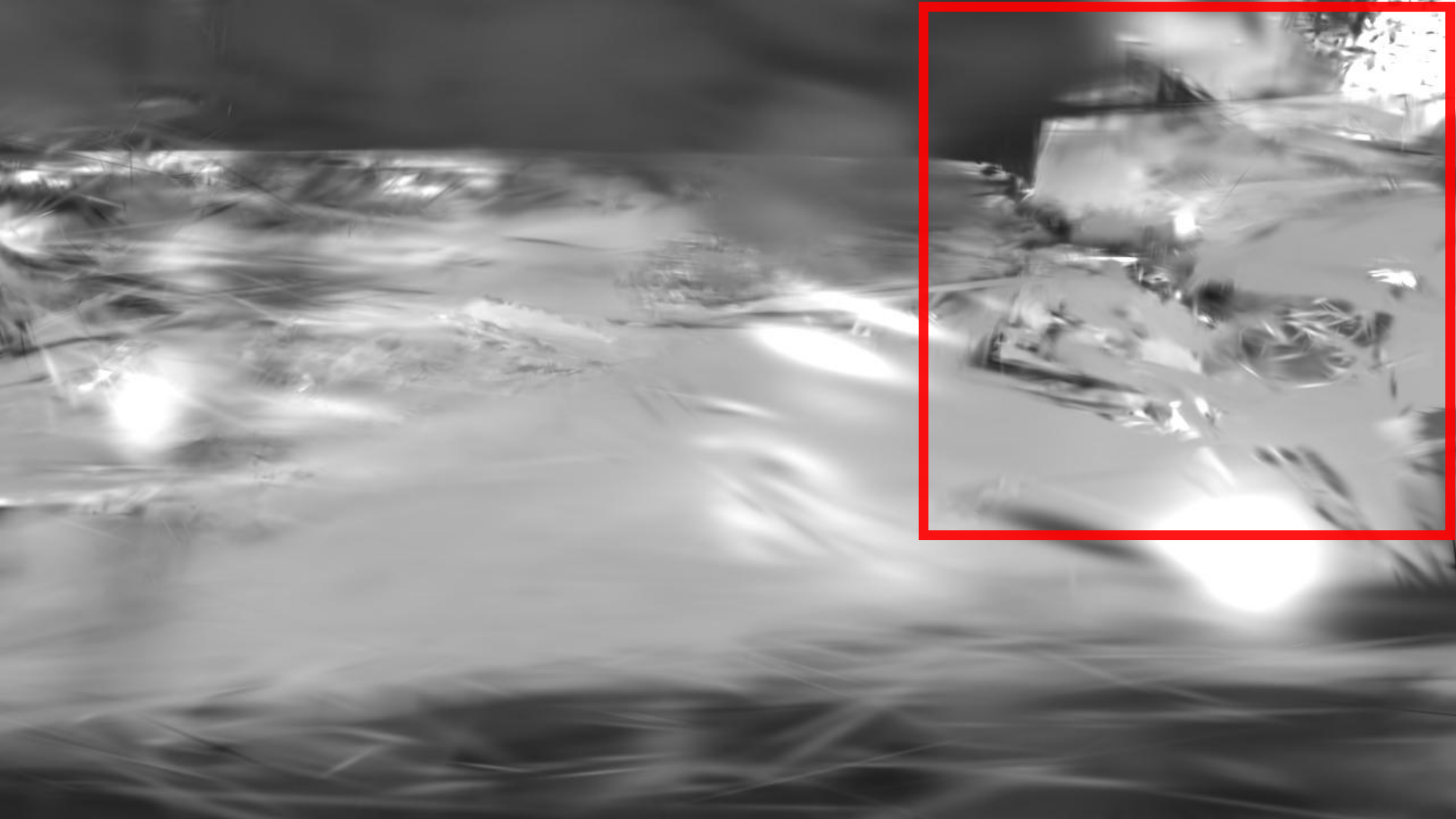} &
\includegraphics[width=0.23\linewidth]{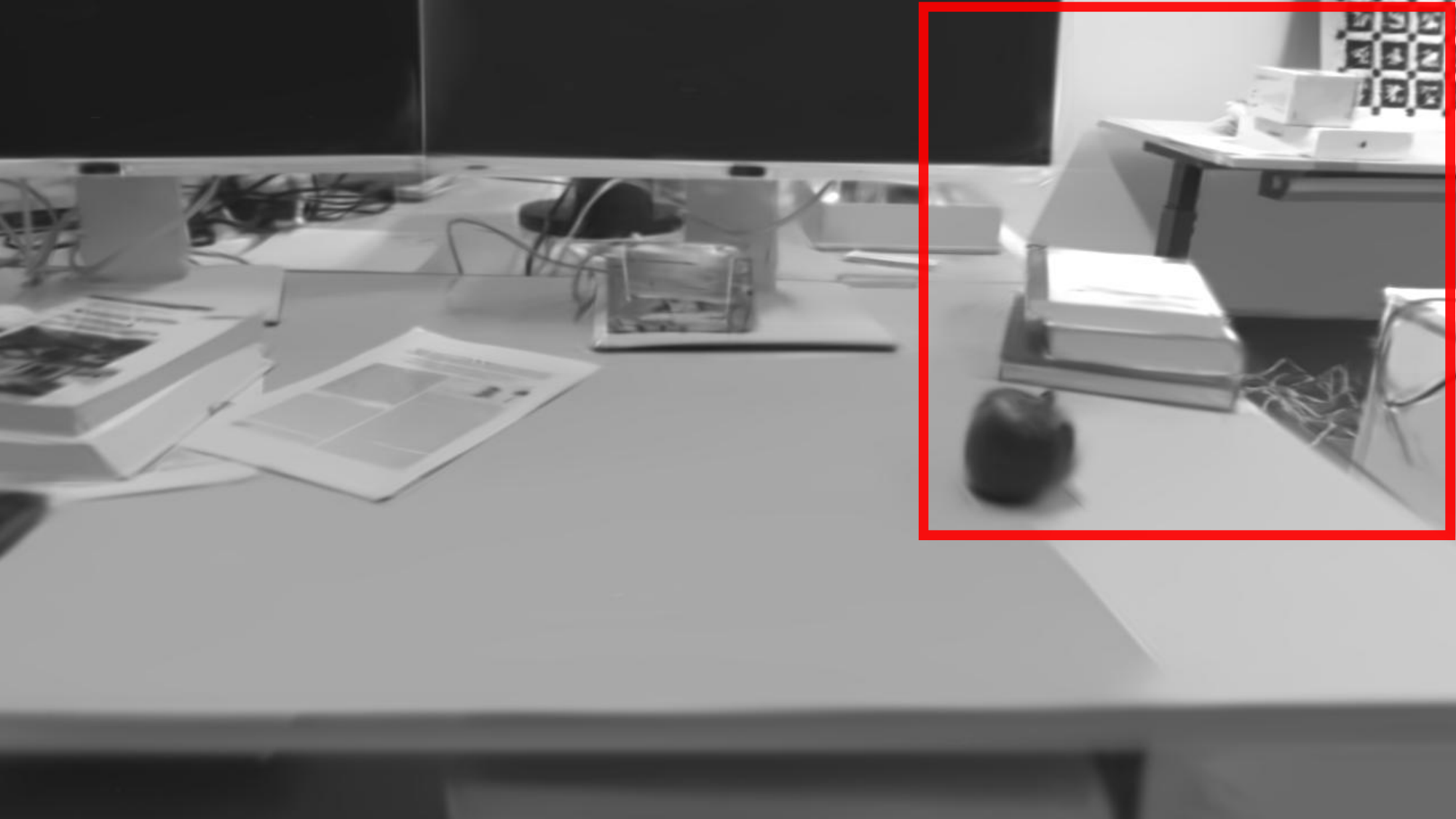} \\
\includegraphics[width=0.23\linewidth]{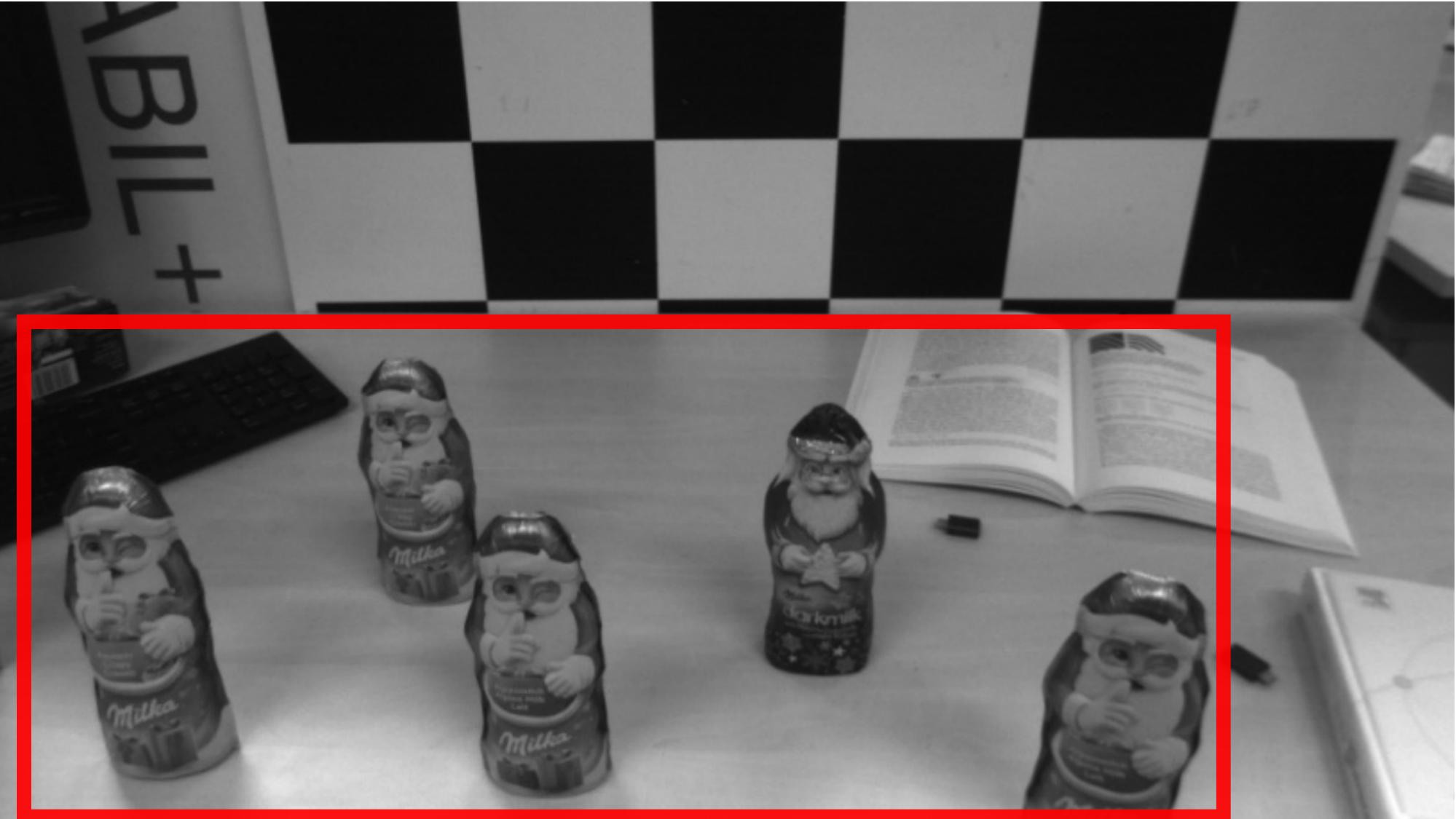} &
\includegraphics[width=0.23\linewidth]{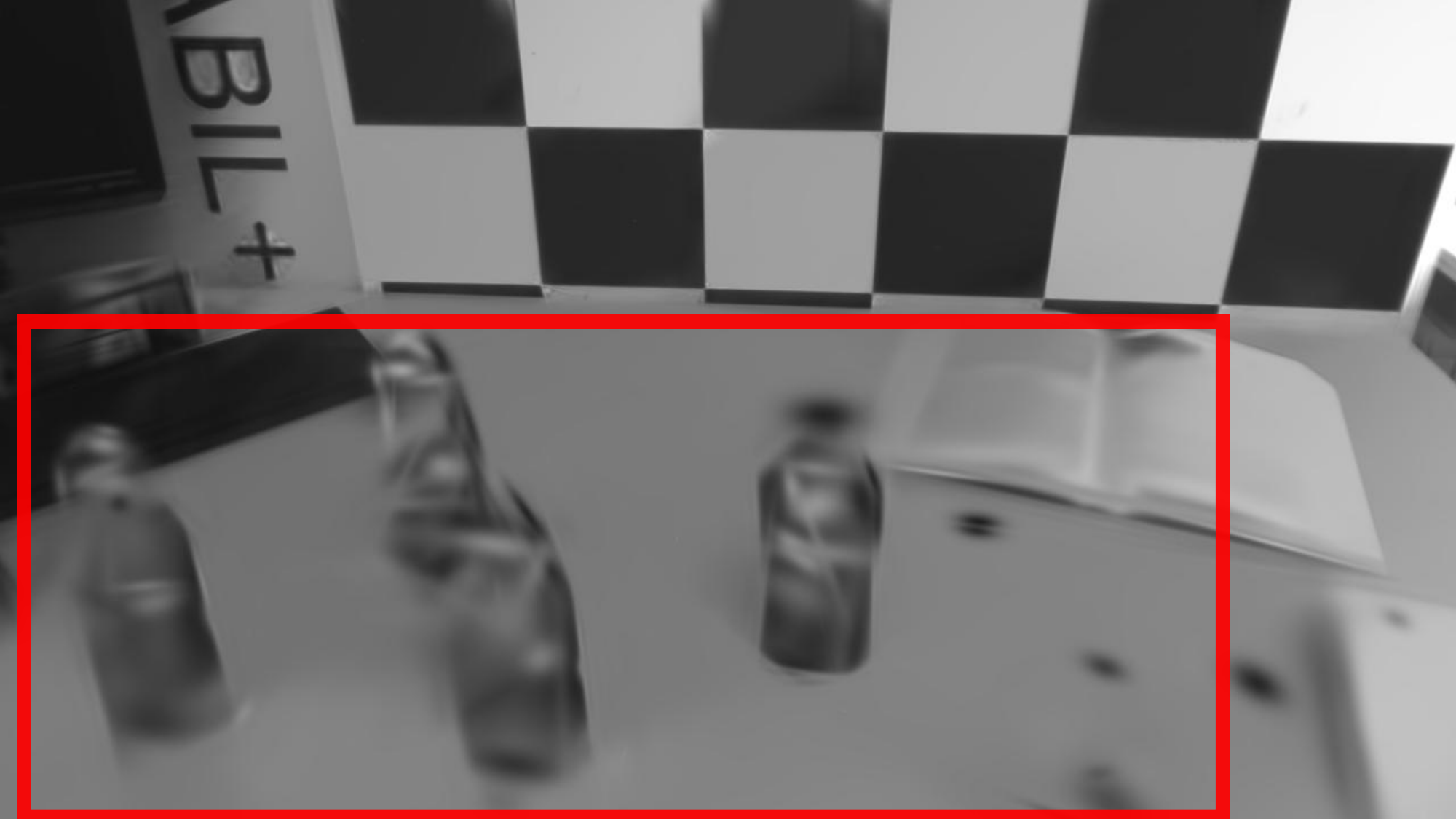} &
\includegraphics[width=0.23\linewidth]{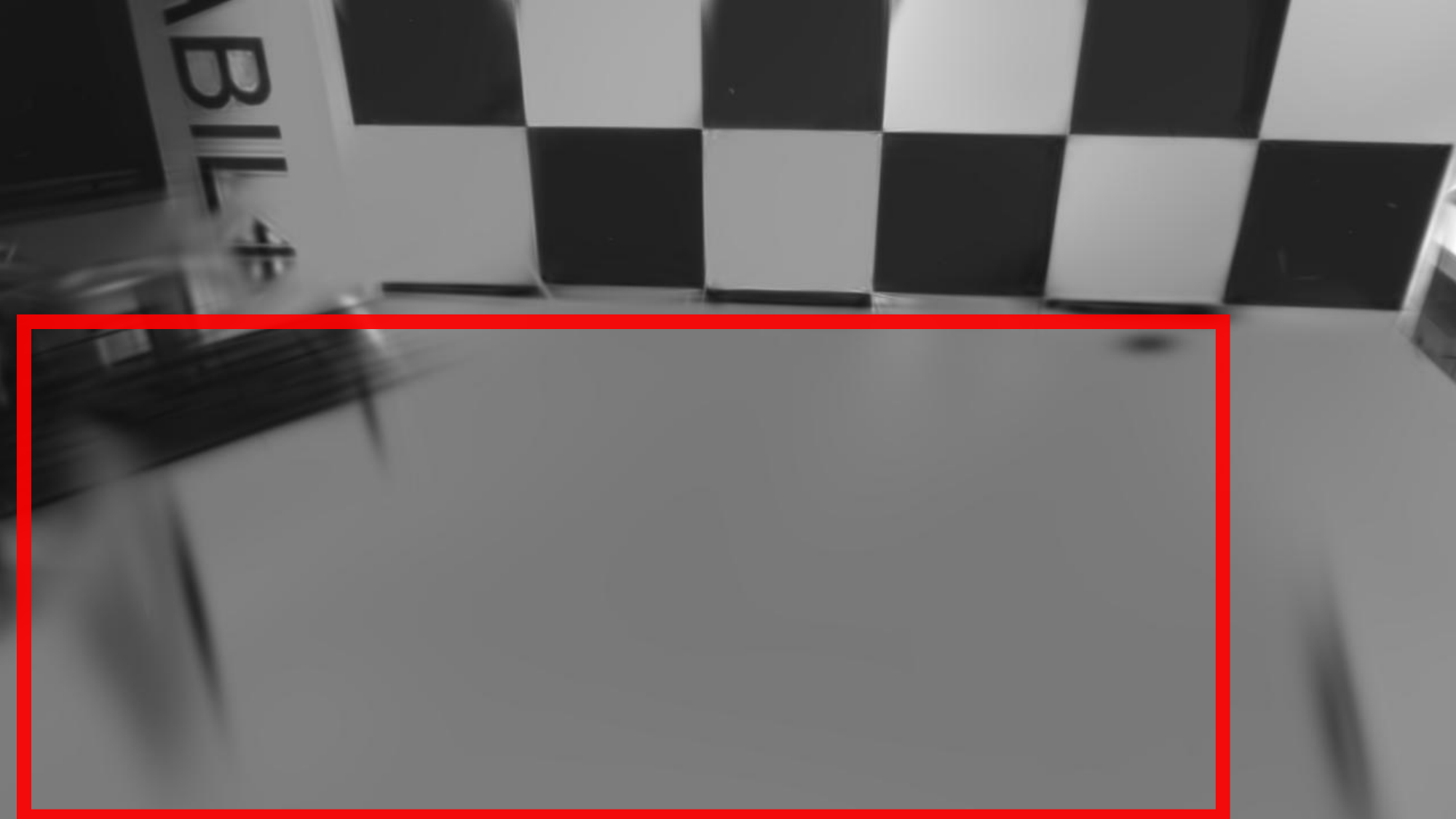} &
\includegraphics[width=0.23\linewidth]{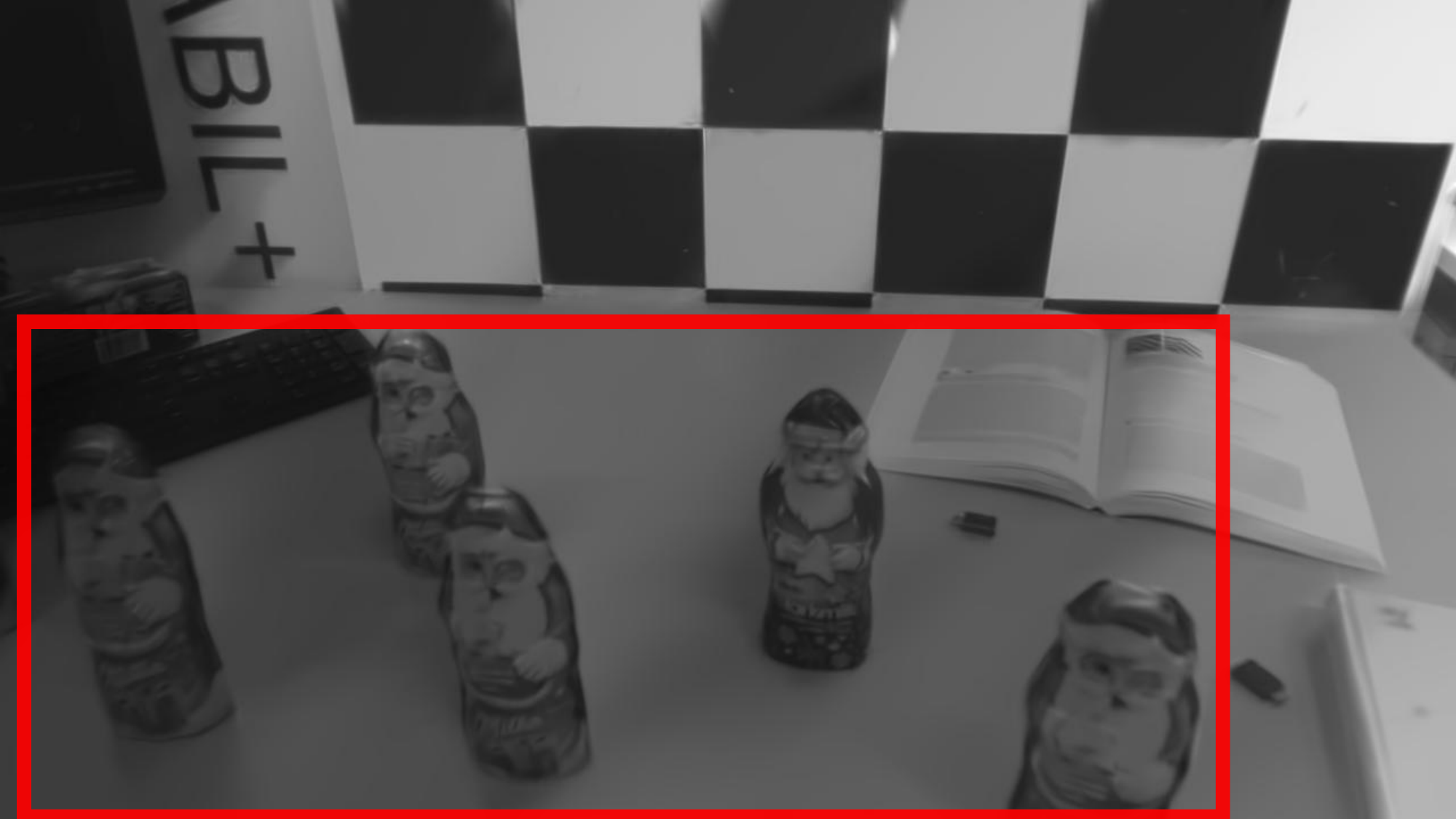} \\
(a) Ground Truth & (b) IncEventGS~\cite{Huang25} & (c) IncEventGS$^\dagger$ & (d) Ours \\
\end{tabular}
\vspace{-2mm}
\caption{\textbf{Impact of trajectory error on reconstruction quality.} 
(a) Ground truth. (b) IncEventGS exhibits multiple failure modes: spatial misalignment causing viewpoint shifts and blurred regions in distant areas beyond initial coverage (top), and objects disappearing (bottom). (c) IncEventGS$^\dagger$ fails to reconstruct scenes due to severe trajectory errors. (d) Our method achieves accurate reconstruction with sharp textures and correct spatial alignment.}
\label{fig:reconstruction_comparison}
\vspace{-4mm}
\end{figure*}

\noindent\textbf{Impact of trajectory error on reconstruction.}
Fig.~\ref{fig:reconstruction_comparison} illustrates how pose accuracy influences the 3D reconstruction quality on the TUM-VIE dataset. IncEventGS$^\dagger$ fails to reconstruct recognizable figurines and produces distorted scenes due to severe trajectory drift. IncEventGS exhibits spatial misalignment with viewpoint shift, blurred distant regions beyond initial coverage, and missing figurines. These degradations occur because depth models estimated from initial frames fail to generalize when cameras explore broader spatial regions. In contrast, our method achieves accurate reconstruction with sharp textures and correct spatial alignment. The proposed edge-guided optimization provides robust geometric constraints by focusing on edge regions, ensuring accurate pose estimation and consistent spatial alignment. Meanwhile, the baseline methods suffer from noisy photometric consistency that accumulates into geometric distortions.


\begin{table}[t]
\caption{\textbf{Ablation study on edge components.}}
\vspace{-2mm}
\centering
\resizebox{\columnwidth}{!}{
\begin{tabular}{l|ccc|c}
\toprule
Method & Edge loss & Edge init & Depth init & ATE (cm) \\
\midrule
IncEventGS~\cite{Huang25} & \ding{55} & \ding{55} & \ding{51} & 0.37 \\
\midrule
IncEventGS$^\dagger$ & \ding{55} & \ding{55} & \ding{55} & 6.62 \\
w/ Edge loss & \ding{51} &  \ding{55} & \ding{55} & 0.50 \\
w/ Edge init & \ding{55} & \ding{51} & \ding{55} & 0.29 \\
Ours & \ding{51} & \ding{51} & \ding{55} & \textbf{0.28} \\
\bottomrule
\end{tabular}}
\label{tab:ablation_module}
\vspace{-4mm}
\end{table}

\begin{figure*}[!ht]
\centering
\begin{tabular}{@{}c@{\hspace{3mm}}c@{}}
\includegraphics[width=0.46\linewidth]{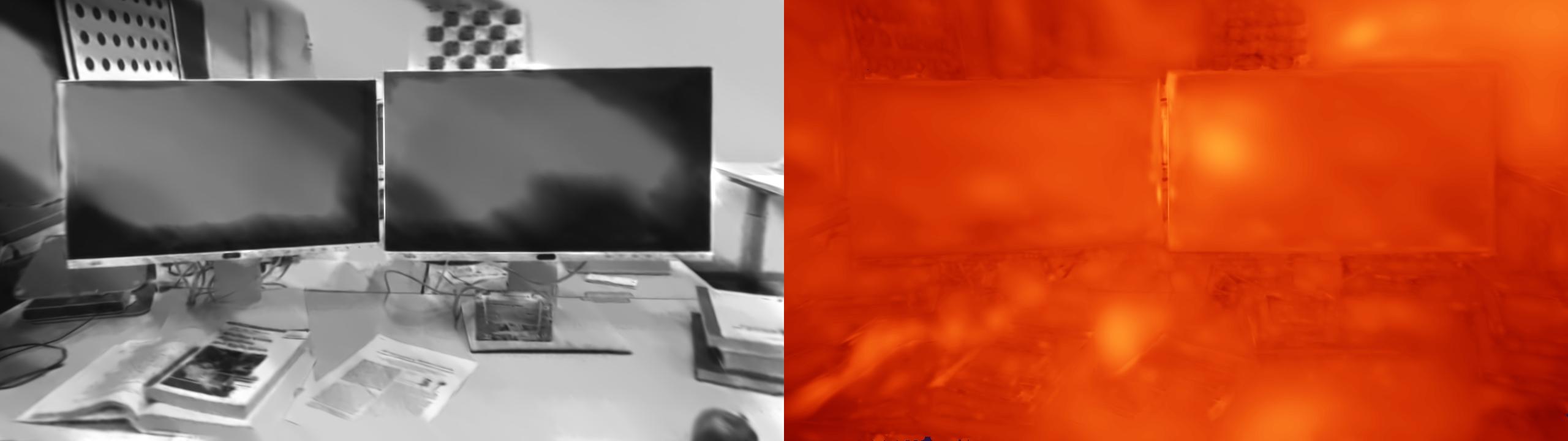} &
\includegraphics[width=0.46\linewidth]{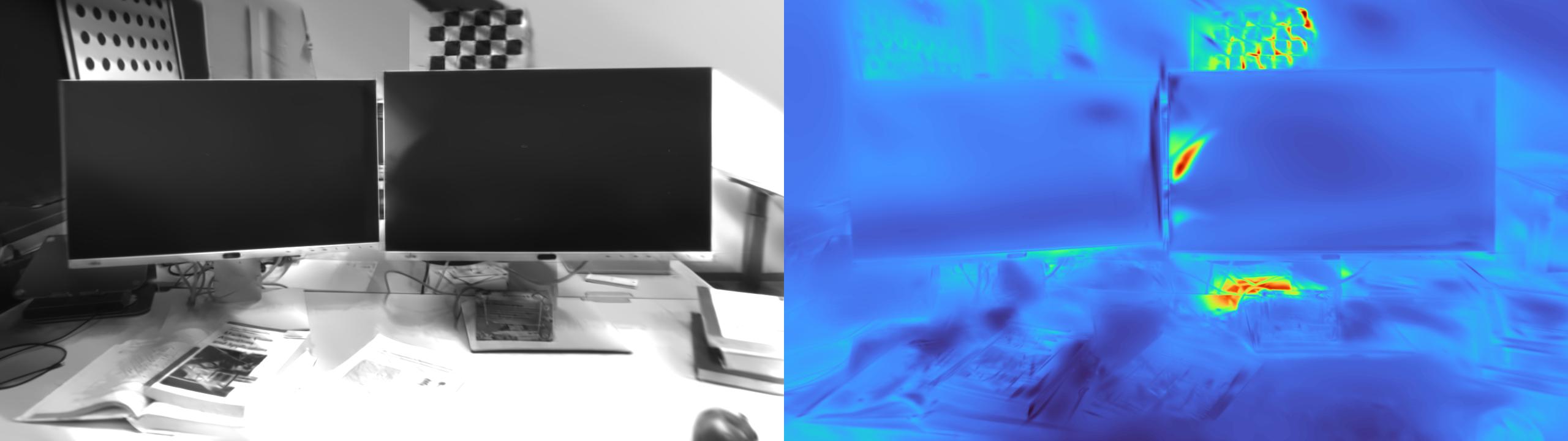} \\
\small (a) IncEventGS$^\dagger$ - 500 iter. & 
\small (b) IncEventGS$^\dagger$ - 3500 iter. \\
\includegraphics[width=0.46\linewidth]{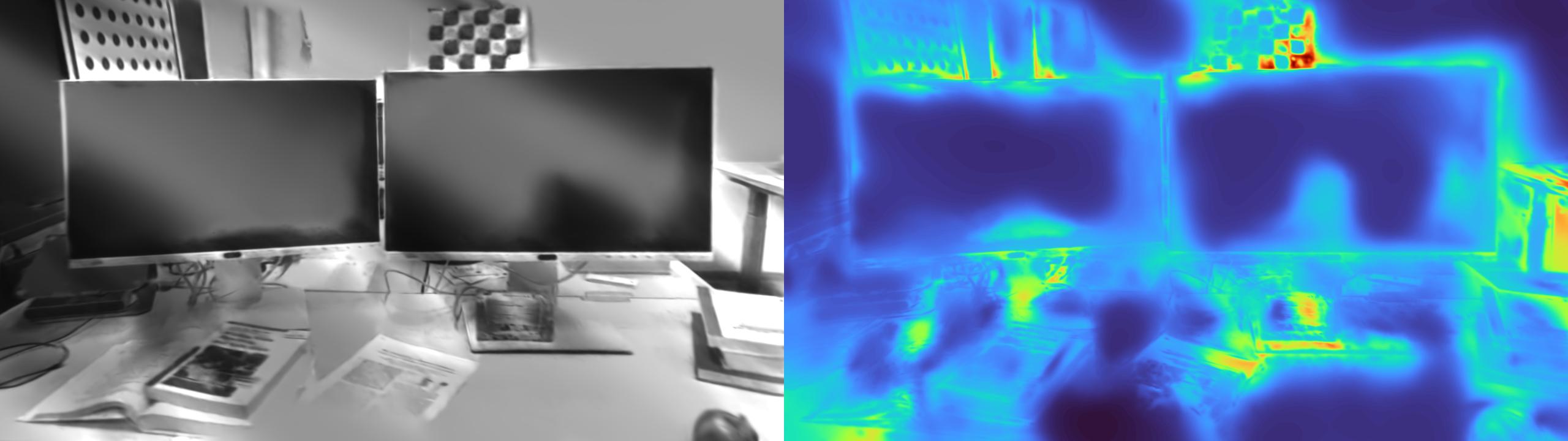} &
\includegraphics[width=0.46\linewidth]{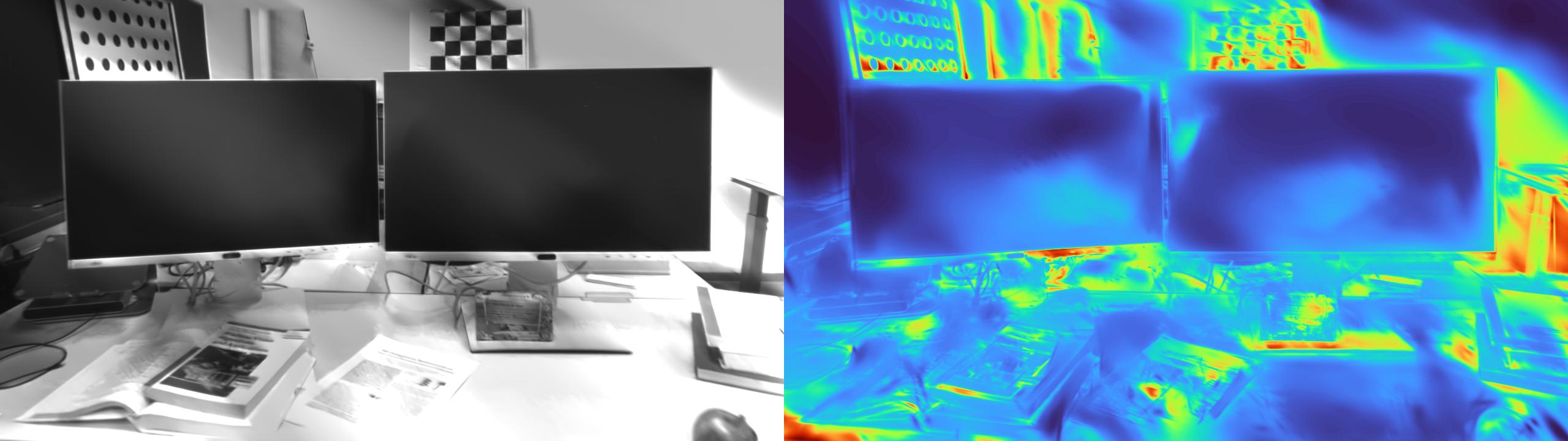} \\
\small (c) Ours ($r_{\text{edge}}=0.0$) - 500 iter. & 
\small (d) Ours ($r_{\text{edge}}=0.0$) - 3500 iter. \\
\end{tabular}
\vspace{-2mm}
\caption{\textbf{Effect of edge-guided loss.} 
Baseline (top) vs. ours with edge loss but random initialization (bottom) at early and final training stages. Each pair shows rendered image (left) and depth map (right). Depth maps show that our edge-guided loss enables faster convergence.}
\label{fig:ablation_edge_loss}
\vspace{-2mm}
\end{figure*}

\begin{figure*}[!ht]
\centering
\begin{tabular}{@{}c@{\hspace{1mm}}c@{\hspace{1mm}}c@{\hspace{1mm}}c@{\hspace{1mm}}c@{}}
\includegraphics[width=0.19\linewidth]{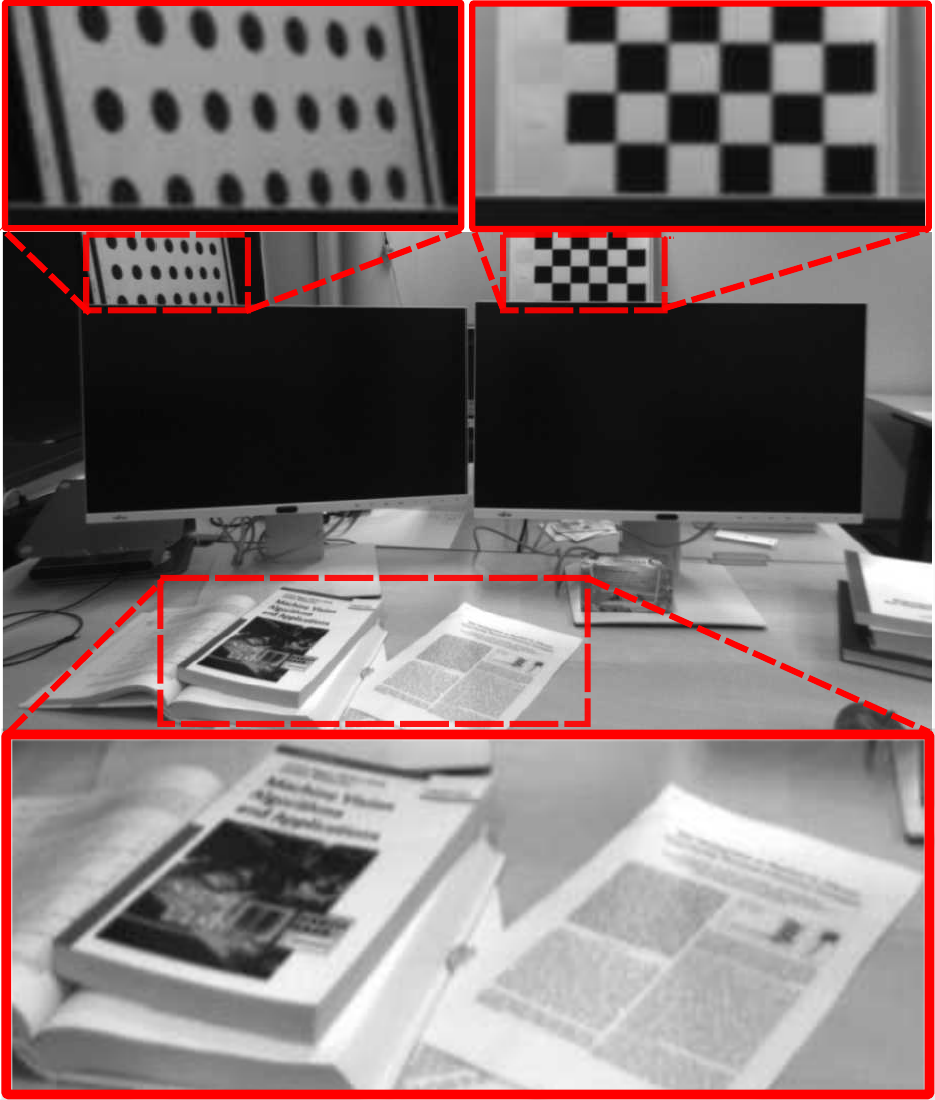} &
\includegraphics[width=0.18\linewidth]{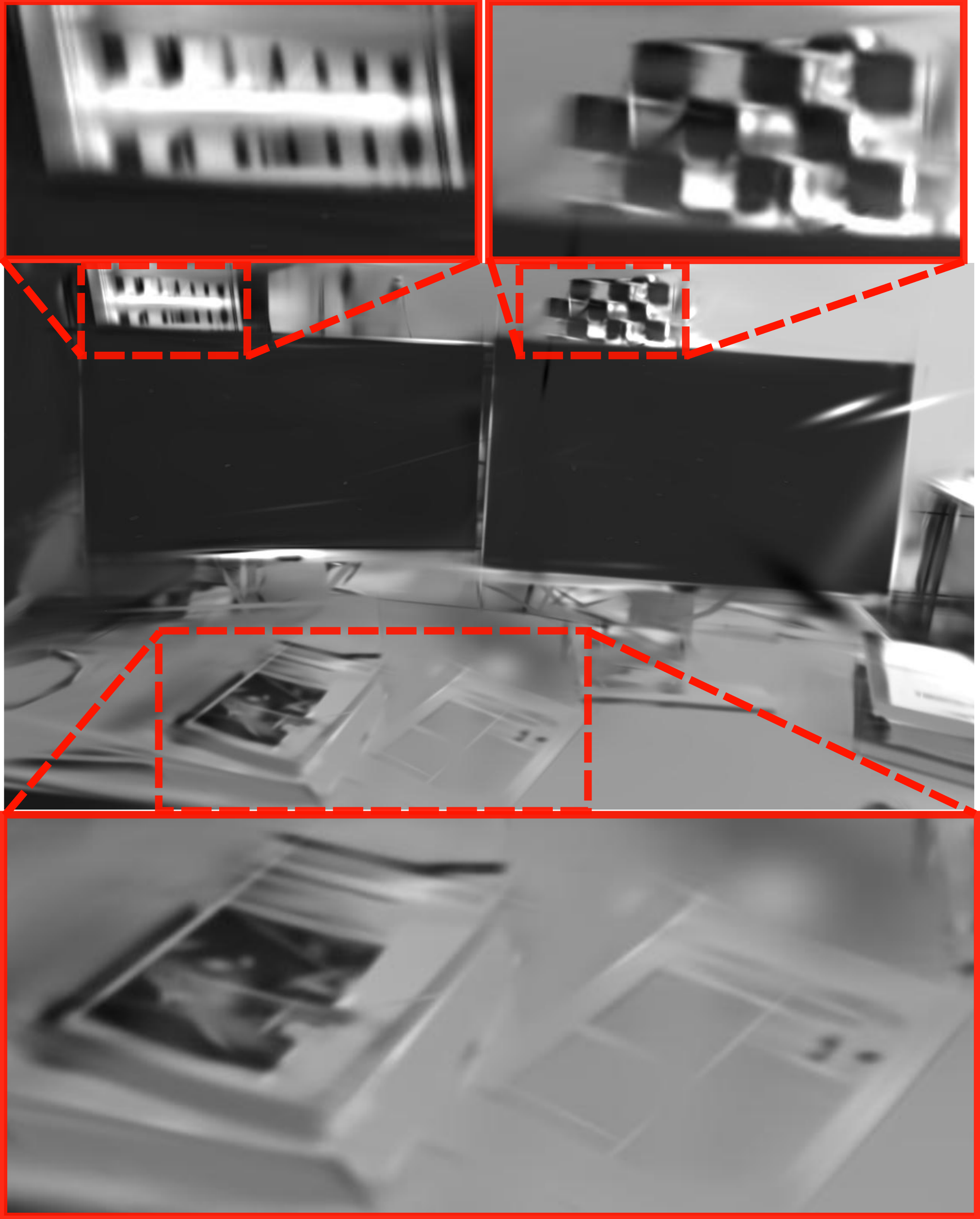} &
\includegraphics[width=0.18\linewidth]{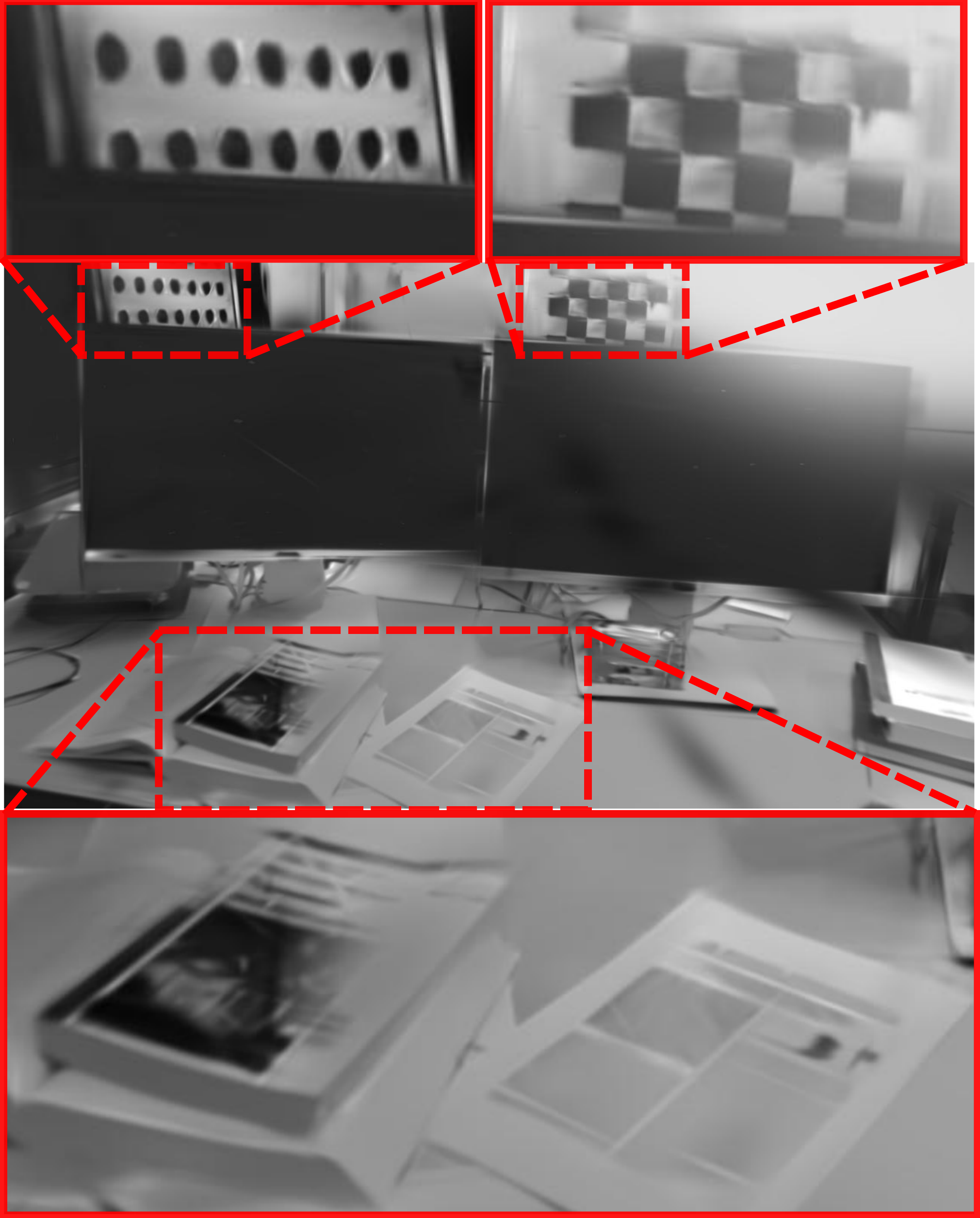} &
\includegraphics[width=0.18\linewidth]{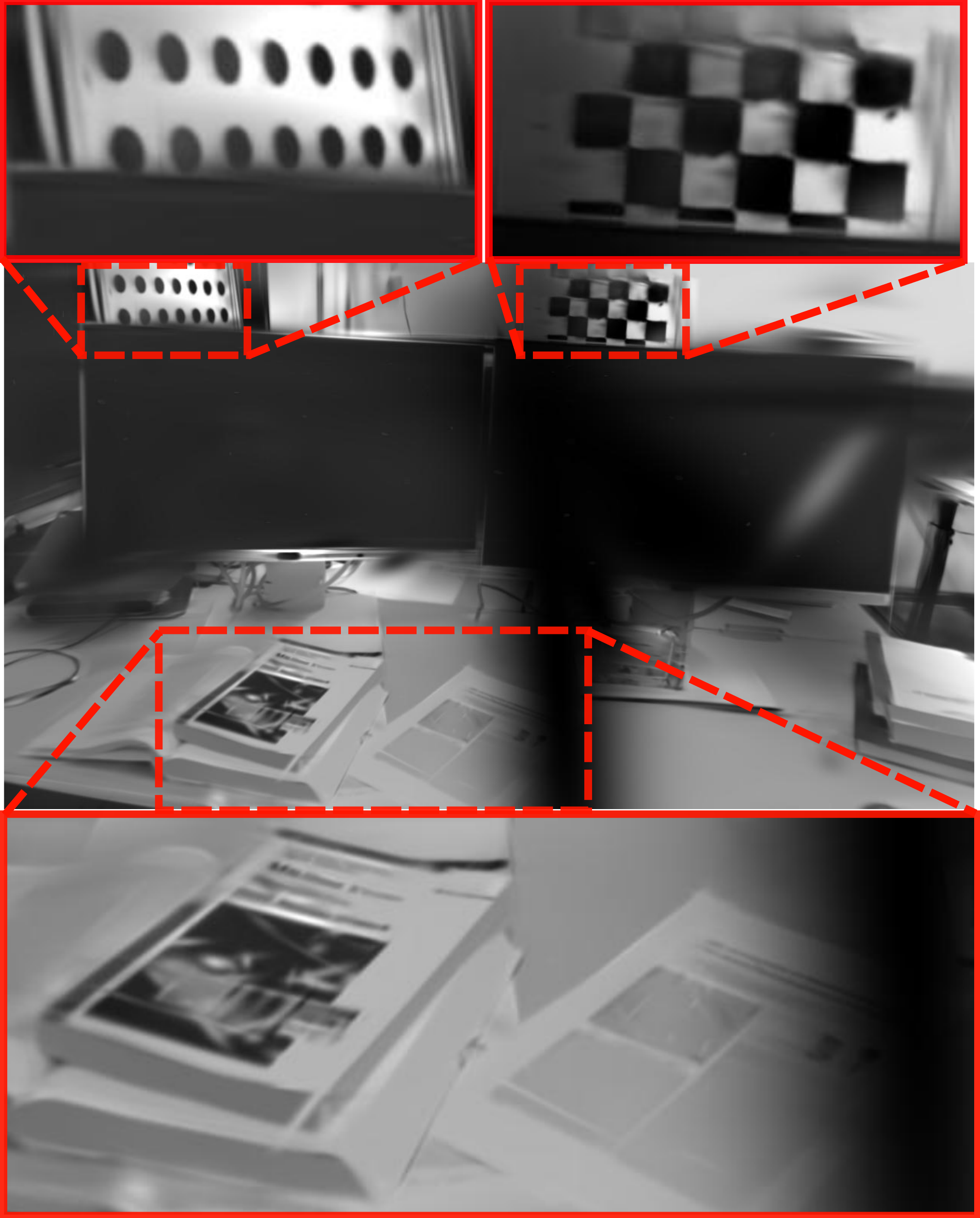} &
\includegraphics[width=0.18\linewidth]{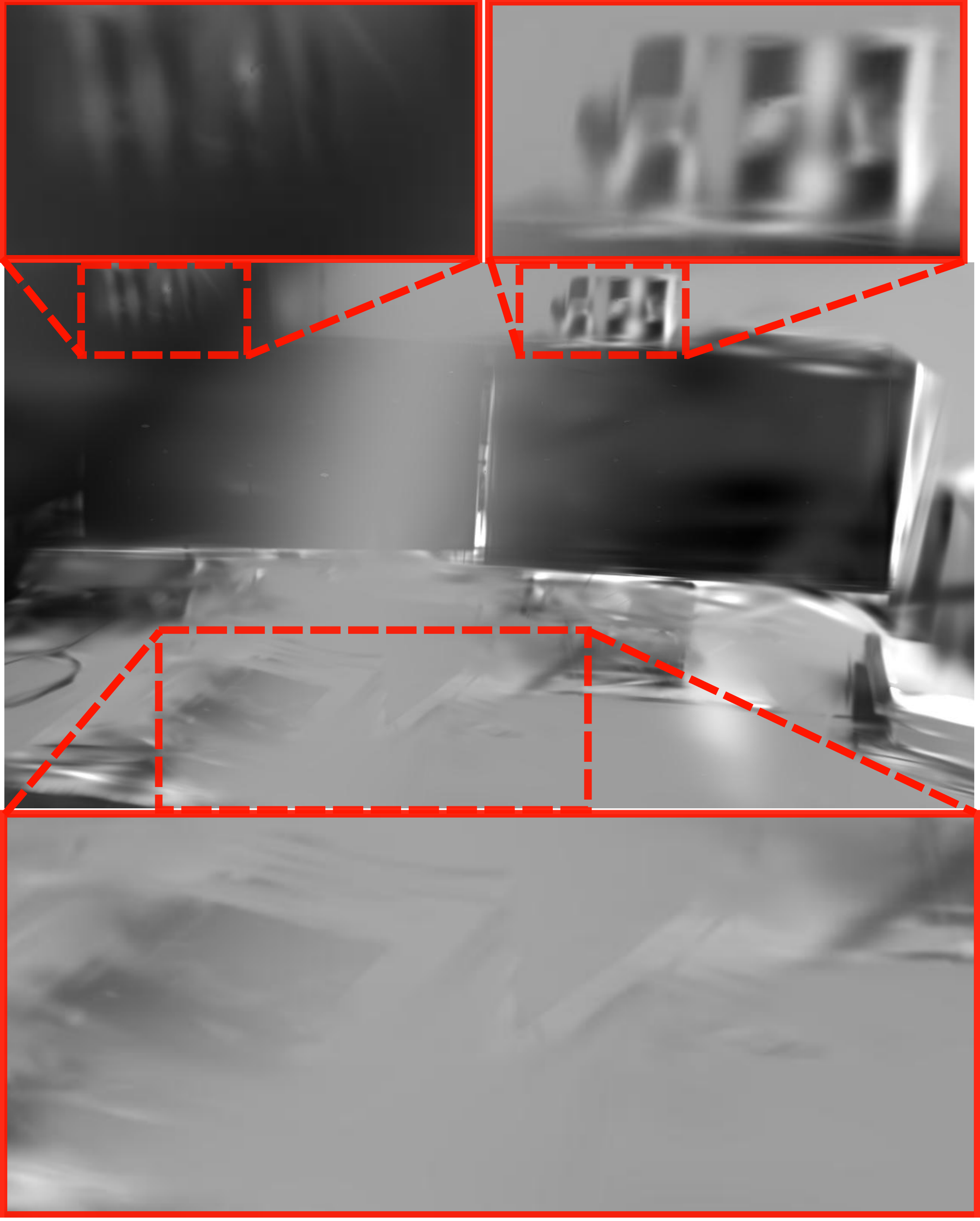} \\
\small (a) Ground Truth &
\small (b) $r_{\text{edge}} = 0.0$ & 
\small (c) $r_{\text{edge}} = 0.3$ & 
\small (d) $r_{\text{edge}} = 0.5$ & 
\small (e) $r_{\text{edge}} = 0.7$ \\
\end{tabular}
\vspace{-2mm}
\caption{\textbf{Effect of edge ratio on reconstruction quality.} 
Red boxes highlight comparison regions. (a) Ground truth. (b) Without edge guidance, fine details such as dot patterns on the back wall are lost. (c-d) Moderate edge ratios achieve optimal reconstruction with well-preserved details. (e) Excessive edge emphasis causes reconstruction degradation due to insufficient surface coverage.}
\label{fig:edge_ratio_eval}
\vspace{-2mm}
\end{figure*}

\subsection{Ablation study}

\begin{table}[t]
\caption{\textbf{Ablation study on edge ratio~($r_\text{edge}$).}}
\centering
\small
\setlength{\tabcolsep}{4pt}
\resizebox{0.42\textwidth}{!}{
\begin{tabular}{c|cccccc}
\toprule
\textbf{Edge Ratio ($r_\text{edge}$)} & 0.0 & 0.1 & 0.3 & 0.5 & 0.7 & 1.0 \\
\midrule
\textbf{ATE (cm)} & 5.68 & 0.41 & 0.40 & 1.78 & 5.53 & 11.93  \\
\bottomrule
\end{tabular}}
\label{tab:ablation_edge_ratio}
\vspace{-4mm}
\end{table}

\textbf{Component-wise ablation.}
To validate the contribution of each proposed component, we conduct component-wise ablation experiments. ~\cref{tab:ablation_module} shows the progressive improvement when adding our components to IncEventGS$^\dagger$. Fig.~\ref{fig:ablation_edge_loss} shows 3D reconstruction with initialized 3D Gaussians at early and final stages of initialization. Our edge-guided loss spatially weights reconstruction error by edge confidence, enabling rapid structure establishment and substantially clearer boundaries at convergence. The loss amplifies gradients at edge locations where depth discontinuities exist, providing stronger supervision for both camera pose and 3D Gaussian parameters. Note that both IncEventGS$^\dagger$ and our method show inaccurate depth estimation in surface regions, as event cameras generate sparse responses in smooth, texture-less regions. This characteristic limits the quality of 3D Gaussians in non-edge regions. However, our edge-guided approach demonstrates clear improvement in geometrically salient regions, which are more critical for accurate pose estimation and scene understanding.

\noindent\textbf{Ablation on parameters settings.}
Fig.~\ref{fig:edge_ratio_eval} and Tab.~\ref{tab:ablation_edge_ratio} demonstrate the impact of edge ratio $r_{\text{edge}}$. The edge ratio $r_{\text{edge}}$ controls the balance between geometric constraints and surface coverage. Without edge initialization, the system experiences trajectory drift, leading to loss of details in 3D reconstruction. We find that $r_{\text{edge}} \in [0.1, 0.3]$ provides sufficient edge constraints, resulting in clear features in 3D reconstruction and low trajectory errors. Excessive edge emphasis under-represents smooth surfaces, leading to incorrect trajectory estimation. While edges provide strong geometric constraints, the lack of information about smooth surfaces makes the difference between the rendered image and the input event map larger, leading to photometric loss dominating in non-edge regions. For detailed analysis of other parameters, please refer to the supplementary material.



\section{Conclusion}
We presented E2EGS, an edge-guided framework for pose-free 3D reconstruction that relies solely on event data. We extract edge maps containing geometric information via temporal coherence analysis and leverage them for structure-aware initialization and optimization. E2EGS achieves superior trajectory estimation and competitive quality in 3D reconstruction. In experiments with long sequences, our proposed method significantly outperforms existing random and depth-based initialization methods, with the effect particularly pronounced when the camera captures new regions beyond the initial image area. This demonstrates the effectiveness of the edge-guided method and expands the domain of completely pose-free event-based 3D reconstruction.


\noindent\textbf{Limitations and future work.} While our method achieves strong results, there is room for improvement. The optimal strategy for edge extraction from noisy event streams remains an open problem. Our current strategy prioritizes reliability over completeness, potentially affecting reconstruction in regions with weak geometric structures. Developing adaptive methods that adjust edge extraction parameters based on local event statistics could improve reconstruction quality in such challenging scenarios.
\clearpage
\putbib[main]
\end{bibunit}


\end{document}